\documentclass[11pt]{article}

\usepackage[preprint]{acl}

\usepackage{times}
\usepackage{latexsym}

\usepackage[T1]{fontenc}

\usepackage[utf8]{inputenc}

\usepackage{microtype}

\usepackage{inconsolata}

\usepackage{graphicx}

\usepackage{microtype}
\usepackage{graphicx}
\usepackage{subfigure}
\usepackage{booktabs} 
\usepackage{longtable}
\usepackage{multirow}
\usepackage{makecell}
\usepackage[ruled]{algorithm2e}
\usepackage{tabularx}
\usepackage{array}
\usepackage{placeins}
\usepackage{wrapfig}
\usepackage{adjustbox}

\usepackage{amsmath}
\usepackage{amssymb}
\usepackage{mathtools}
\usepackage{amsthm}

\theoremstyle{plain}

\theoremstyle{definition}

\theoremstyle{remark}

\usepackage[textsize=tiny]{todonotes}

\title{GUIDE: Guiding Internal Evidence with Language Instructions}

\newcounter{eqfn}
\newcounter{secfn}
\newcounter{corrfn}

\author{%
Soyeon Caren Han$^{1}$\thanks{Equal contribution.}\setcounter{eqfn}{\value{footnote}}\thanks{Corresponding authors: Soyeon Caren Han \texttt{caren.han@unimelb.edu.au}, Kyungreem Han \texttt{khan@kist.re.kr}}\setcounter{corrfn}{\value{footnote}}, 
Hyunsuk Chung$^{1}$\footnotemark[\value{eqfn}], 
Jinwoo Kim$^{1}$\thanks{These authors contributed equally as second authors.}\setcounter{secfn}{\value{footnote}}, 
Seungyeon Ji$^{2,3}$\footnotemark[\value{secfn}],
Kyungreem Han$^{2,4}$\footnotemark[\value{corrfn}]\\
\mdseries
$^1$School of Computing and Information Systems, The University of Melbourne, Melbourne, Australia \\
$^2$Brain Science Institute, Korea Institute of Science and Technology, Seoul, Republic of Korea \\
$^3$Department of Computer Science and Engineering, Korea University, Seoul, Republic of Korea \\
$^4$Division of Bio-Medical Science \& Technology, University of Science and Technology KIST School, \\Seoul, Republic of Korea \\
\texttt{\{caren.han, hyunsuk.chung.1, jinwoo.kim.1\}@unimelb.edu.au} \\
  \texttt{\{syji, khan\}@kist.re.kr}
}

\begin{document}
\maketitle

\begin{abstract}
Large multimodal models follow instructions about what to generate, but not necessarily about what evidence to rely on. Hence, models may continue to depend on shortcut-associated cues even when instructions suggest otherwise.
We introduce GUIDE, a framework for controlling internal evidence usage through language instructions. GUIDE combines grouped parameter-efficient adaptation with instruction-conditioned gating to modulate multimodal evidence pathways during reasoning and generation.
We further introduce a pathway-level evaluation framework that characterizes instruction-conditioned evidence modulation through reliance sensitivity, controlled perturbation analysis, pathway modulation, and autoregressive decoding dynamics.
Across multimodal reasoning, classification, and generation, GUIDE induces structured and instruction-aligned redistribution of evidence reliance while largely preserving task behavior.
Experiments on GQA, TextVQA, MM-IMDb, CREMA-D, RAVDESS, and Flickr30K show that GUIDE improves robustness under targeted evidence perturbations and enables controllable modulation across diverse multimodal settings.
This suggests that multimodal instruction following can extend beyond output control toward regulating how different evidence sources contribute to model predictions.
\end{abstract}

\section{Introduction}
Large multimodal models can often follow instructions about what to generate, but not necessarily about what evidence to rely on.
For example, a model instructed to \textit{``focus on actions''} may depend on appearance, demographic, or contextual cues while producing the same answer. Similarly, instructions such as \textit{``ignore appearance''} or \textit{``rely on textual evidence''} may change model outputs without systematically changing the evidence sources that contribute to the prediction.
This limitation is particularly important in multimodal settings, where multiple evidence sources can independently support the same task outcome.
In tasks such as GQA and TextVQA, models can rely on object relationships, OCR signals, visual attributes, or background context to produce identical answers.
Moreover, shortcut-associated cues are often highly predictive under correlation-driven learning, encouraging models to exploit spurious evidence even when instructions suggest otherwise~\cite{geirhos_shortcut_learning, shortcut_llm, spurious_benchmm}.
As a result, controlling outputs alone does not necessarily imply control over how information contributes to model predictions.
Recent progress in multimodal instruction following has primarily focused on prompt engineering, instruction tuning, and controllable generation.
While these approaches improve output controllability, they provide limited mechanisms for explicitly regulating instruction-conditioned evidence reliance~\cite{attention_control, visual_prompting, turner2023steering, zou2023representation}.

\begin{figure}[t]
    \centering
    \includegraphics[width=\columnwidth]{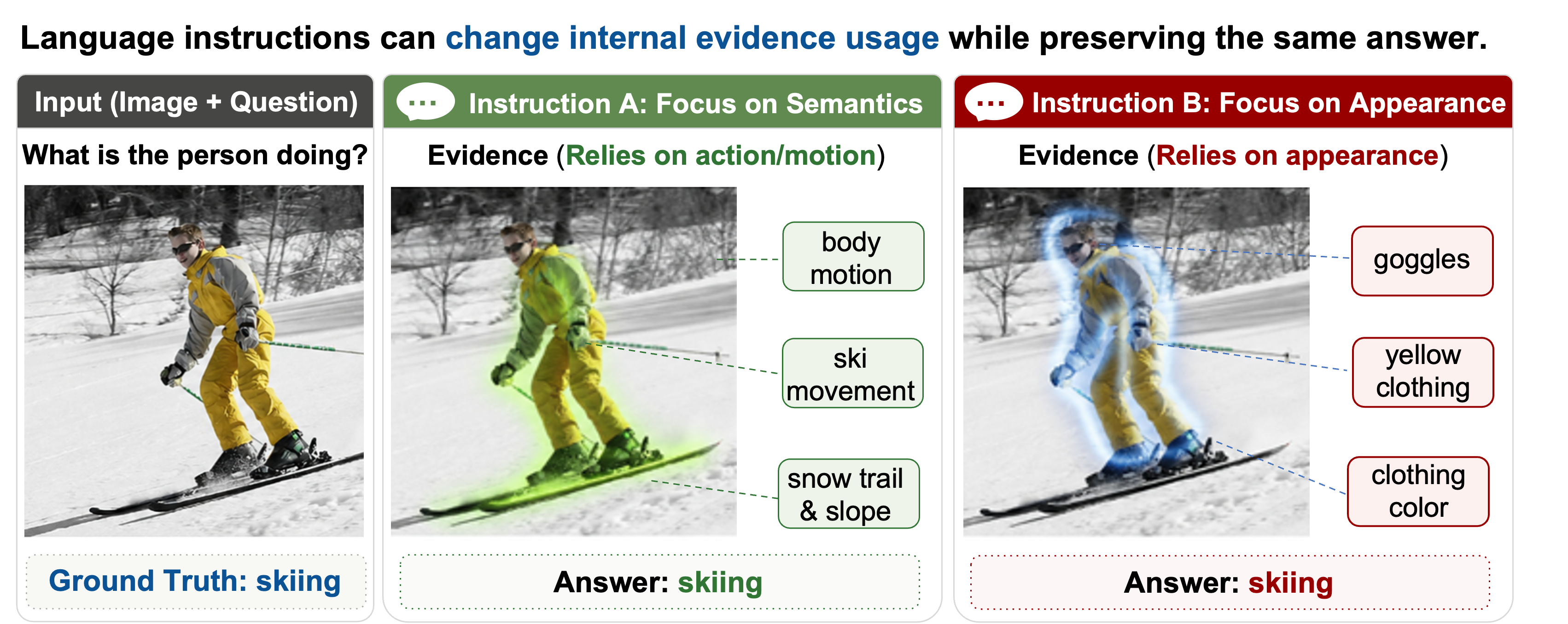}
    \vspace{-1.3em}
    \caption{Motivation of GUIDE. 
Given the same input image and question, different language instructions encourage reliance on different internal evidence pathways while producing the same answer. 
Semantic-focused instructions emphasize action and interaction cues, whereas appearance-focused instructions increase reliance on visual attribute evidence.}
\vspace{-1.1em}
    \label{fig:FiLoRA}
\end{figure}

An important open problem remains: can language instructions systematically regulate the evidence pathways used by multimodal models?
To address this problem, we introduce \textbf{GUIDE}, a framework for controlling multimodal evidence usage with language instructions.
GUIDE combines grouped parameter-efficient adaptation with instruction-conditioned gating to dynamically modulate task-relevant and taxonomy-aligned evidence pathways.
Rather than modifying task objectives or input representations, GUIDE operates through grouped internal adaptation pathways to regulate the contribution of multimodal evidence during reasoning and generation.
Importantly, GUIDE enables models to produce similar outputs while exhibiting substantially different pathway activation and functional reliance patterns.
For example, semantic-focused instructions increase reliance on action and interaction pathways, whereas appearance-focused instructions increase reliance on visual attribute pathways.
Similar to this, in TextVQA, instructions that emphasize textual evidence increase reliance on OCR-related pathways while reducing dependence on contextual appearance cues.

To study this behavior, we introduce a pathway-level evaluation framework based on pathway modulation, controlled perturbation analysis, reliance sensitivity, and autoregressive decoding dynamics.
Across multimodal reasoning, classification, and generation tasks, GUIDE induces structured and instruction-aligned redistribution of evidence reliance while largely preserving task behavior.
We evaluate GUIDE on GQA, TextVQA, MM-IMDb, CREMA-D, RAVDESS, and Flickr30K.
Experiments further show improved robustness under targeted evidence perturbations.
Our main contributions are as follows:
\textbf{1)} We introduce GUIDE, a framework for instruction-conditioned control of multimodal evidence usage through grouped adaptation pathways and dynamic gating.
\textbf{2)} We provide pathway-level and perturbation-based evidence that language instructions systematically redistribute model reliance across task-relevant multimodal evidence sources.
\textbf{3)} We demonstrate that instruction-conditioned evidence control generalizes across multimodal reasoning, classification, and generation tasks while preserving task semantics.
Together, these findings suggest that multimodal instruction following can extend beyond output control toward regulating how different evidence sources contribute to model predictions.

\begin{figure*}[t]
    \centering
    \includegraphics[width=\textwidth]{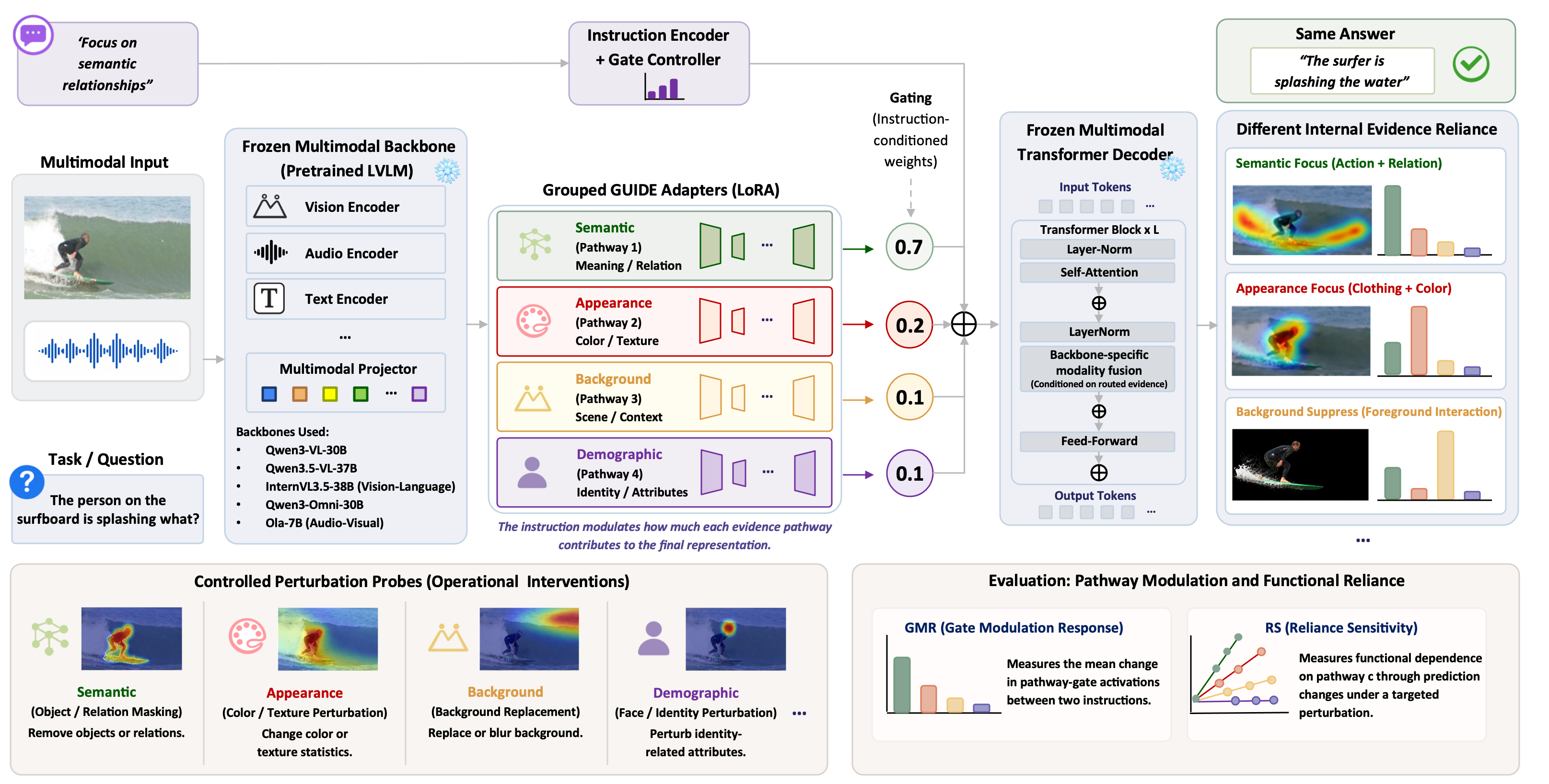}
    \vspace{-1.3em}
    \caption{Overview of GUIDE, an instruction-conditioned evidence routing framework for controllable multimodal reasoning. GUIDE dynamically modulates grouped adaptation pathways corresponding to different evidence categories through instruction-conditioned gating, enabling the same prediction to be produced under different internal evidence reliance patterns. Controlled perturbation probes are used to evaluate both pathway modulation and functional reliance shifts across evidence categories.
}
\vspace{-1em}
    \label{fig:overall_GUIDE}
\end{figure*}

\section{Related Work}

\paragraph{Multimodal Instruction Following.}
Recent multimodal foundation models~\cite{llava_next, instructblip, qwen2_vl, kosmos2} have substantially improved instruction-guided reasoning and controllable vision--language generation.
However, most existing approaches primarily focus on aligning model outputs with user instructions rather than controlling the internal evidence used to produce them.
Methods based on prompt engineering, controllable decoding, attention steering, or visual grounding~\cite{attention_control, visual_prompting, cheng2024spatialrgpt} can influence generated responses or attention distributions, but provide limited explicit control over feature-level evidence reliance during internal computation.
This limitation is particularly important because multimodal models often exploit shortcut-associated cues whenever such correlations are predictive under the training distribution~\cite{geirhos_shortcut_learning, shortcut_llm, spurious_benchmm}.
In multimodal settings, these shortcuts may correspond to appearance attributes, contextual backgrounds, demographic signals, motion patterns, acoustic cues, or OCR artifacts.
Such spurious reliance has been repeatedly observed in multimodal benchmarks including MM-IMDb, CREMA-D, and RAVDESS.

\vspace{-0.5em}
\paragraph{Feature Steering and Representation Control.}
Recent work on feature steering, representation editing, and controllable generation has explored manipulation of latent representations or activations to alter model behavior~\cite{fit, controllingcav, repres_edit, turner2023steering, zou2023representation}.
These approaches demonstrate that model behavior can often be influenced through targeted representation-level interventions.
However, existing methods primarily focus on text-only settings or activation-level manipulation, primarily through post-hoc activation manipulation rather than instruction-conditioned multimodal evidence routing.
They do not explicitly regulate how multimodal evidence is internally selected and combined during reasoning and generation.

\vspace{-0.5em}
\paragraph{Bias Mitigation and Parameter-Efficient Adaptation.}
Multimodal bias mitigation methods~\cite{assume, multimodal-representation-learning, geometric} primarily operate during training to reduce dataset-level spurious correlations.
While effective for improving robustness, these approaches generally do not support user-specified evidence control during inference.
Parameter-efficient fine-tuning methods such as LoRA~\cite{lora}, QLoRA~\cite{qlora}, adapters~\cite{adapter}, FocalLoRA~\cite{focallora}, and mixture-of-LoRA routing approaches~\cite{huynh2025mixlora, gao2025mola} enable efficient adaptation through lightweight parameter updates, but rely on static adaptation pathways without instruction-conditioned modulation of feature-specific computation.

\vspace{-0.5em}
\paragraph{Our Perspective.}
Prior work provides limited mechanisms for translating language instructions into explicit control over internal multimodal evidence usage.
In contrast, GUIDE treats instructions as interventions on evidence reliance itself, enabling language-conditioned modulation of task-relevant multimodal evidence pathways during reasoning and generation. This enables GUIDE to separate output controllability from evidence controllability, allowing similar predictions to arise from different instructed evidence pathways.

\section{GUIDE}
\label{sec:guide}
\vspace{-0.1em}
\subsection{Problem Formulation}

GUIDE addresses the problem of controlling which evidence a multimodal model relies on while preserving task semantics.
Given multimodal input $x$, instruction $I$, and pretrained model $f_{\theta}$, conventional instruction following primarily controls the output: $y=f_{\theta}(x,I)$. However, different evidence sources may independently support the same prediction. For example, a model may answer a question using object relationships, OCR signals, appearance attributes, or contextual background cues.
Our goal is therefore not merely to control outputs, but to modulate the evidence pathways contributing to them.
Specifically, we seek instruction-conditioned evidence modulation such that
$f_{\theta}(x,I_a)\approx f_{\theta}(x,I_b)$
while exhibiting substantially different pathway activation and functional reliance patterns under instructions $I_a$ and $I_b$.

\vspace{-0.1em}
\subsection{Taxonomy-Aligned Feature Pathways}
GUIDE treats multimodal reasoning as composition over heterogeneous evidence sources. To enable controllable evidence modulation, we organize internal adaptation pathways into task-relevant groups associated with different forms of multimodal evidence. These include semantic interactions, visual appearance, environmental context, motion dynamics, acoustic signals, demographic cues, and OCR/textual grounding information depending on the benchmark.
While the routing mechanism is shared across tasks, the pathway taxonomy is adapted to the dominant evidence structure of each dataset.
High-level evidence families may be instantiated as multiple fine-grained operational routing groups; for example, GQA decomposes semantic evidence into object, relation, and action groups, while appearance evidence includes an attribute group.
Each operational pathway corresponds to a dedicated grouped low-rank adaptation branch attached to shared transformer projections.
Pathway assignments follow benchmark-specific evidence taxonomies and weak evidence-localization heuristics described in Appendix~\ref{app:taxonomy_construction}.

\vspace{-0.1em}
\subsection{Grouped LoRA Adaptation}
GUIDE implements instruction-conditioned evidence modulation through grouped parameter-efficient adaptation.
Given pretrained weight matrix $W$, standard LoRA introduces a low-rank update
$W'=W+BA$, where $A$ and $B$ are trainable low-rank matrices.
Instead of using a single shared adaptation pathway, GUIDE decomposes the update into multiple taxonomy-aligned adaptation groups:
\vspace{-1em}
\[
W'=W+\sum_{c=1}^{C} g_c(I,h_t)\,B_cA_c,
\]
where $c$ indexes operational routing groups, $B_cA_c$ denotes the corresponding low-rank adaptation, and $g_c(I,h_t)$ is an instruction-conditioned gate predicted from the instruction representation and current model state.
The gating mechanism dynamically modulates the contribution of each grouped adaptation pathway during reasoning and generation.
Unlike static parameter-efficient adaptation, this formulation allows language instructions to selectively amplify or suppress specific pathways at inference time.
GUIDE does not require explicit disentanglement of internal representations; instead, grouped adaptation provides a structured intervention space for modulating evidence reliance while preserving the underlying task objective.

\vspace{-0.1em}
\subsection{Instruction-Conditioned Gating}
To regulate evidence usage, GUIDE predicts pathway-specific gates conditioned on both the instruction representation and the current model state.
Given instruction embedding $e_I$ and hidden representation $h_t$, the gating network computes:
\vspace{-0.5em}
\[
g(I,h_t)=\sigma\!\left(W_g[e_I;h_t]\right),
\]
where $g(I,h_t)\in[0,1]^C$ contains independently activated gates over $C$ taxonomy-aligned routing groups.
Unlike mutually exclusive routing, these gates allow multiple evidence pathways to be active simultaneously while enabling instructions to selectively increase or suppress individual pathway contributions.
Because gating depends on both the instruction and the evolving model state, pathway activations can change throughout reasoning and generation.
For autoregressive tasks, gates are computed at each decoding step and shared across grouped adaptation modules within the corresponding transformer layer.
GUIDE is trained using the original task objective while jointly optimizing grouped adaptation pathways and instruction-conditioned gating networks.
Additional training details and regularization objectives are provided in Appendix~\ref{app:training_objective}.

\vspace{-0.1em}
\subsection{Reliance Modulation Objective}
To evaluate whether language instructions alter evidence usage, we use two complementary measures: Gate Modulation Response (GMR) and Reliance Sensitivity (RS).
\vspace{-0.3em}
\paragraph{Gate Modulation Response (GMR).}
GMR measures how strongly pathway activations respond to instruction changes.
Given two instructions $I_a$ and $I_b$, we compute:
\vspace{-1em}
\[
\mathrm{GMR}(I_a,I_b)
=
\frac{1}{C}
\sum_{c=1}^{C}
|g_c(I_a)-g_c(I_b)|.
\]
For autoregressive tasks, pathway differences are averaged across decoding steps and gated transformer layers.
For classification tasks, activations are aggregated from the final prediction representation.
Additional implementation details are provided in Appendix~\ref{app:gmr_details}.
Higher GMR indicates stronger instruction-conditioned pathway modulation.

\paragraph{Reliance Sensitivity (RS).}
While GMR measures pathway modulation, it does not necessarily imply functional influence on predictions.
We therefore use RS to measure how strongly model predictions depend on evidence associated with a given pathway.
For pathway group $c$,
\vspace{-0.5em}
\[
\mathrm{RS}_c(I)
=
\mathbb{E}
\left[
\left|
\log p(y|x,I)
-
\log p(y|x,\Delta_c,I)
\right|
\right],
\]
where $\Delta_c$ denotes a controlled perturbation targeting evidence associated with pathway $c$.
Perturbations include modality-specific masking, corruption, replacement, and suppression operations designed to selectively degrade the targeted evidence source while preserving overall task solvability.
Detailed protocols are provided in Table~\ref{tab:perturbation_protocols}.

When comparing two instruction conditions, we summarize the change in functional reliance as
\vspace{-0.5em}
\[
\Delta\mathrm{RS}(I_a,I_b)
=
\frac{1}{C}
\sum_{c=1}^{C}
\left|
\mathrm{RS}_c(I_a)-\mathrm{RS}_c(I_b)
\right|.
\]
Higher $\mathrm{RS}_c$ indicates greater functional dependence on pathway $c$, while higher $\Delta\mathrm{RS}$ indicates stronger redistribution of functional reliance between instructions.
Together, GMR and RS characterize complementary aspects of instruction-conditioned evidence modulation: GMR measures changes in pathway activation, whereas RS measures corresponding changes in functional dependence.


\section{Evaluation Setup}
\vspace{-0.5em}
We evaluate whether GUIDE enables controllable modulation of multimodal evidence usage while preserving task semantics.
To assess both controllability and generalization, we consider multimodal reasoning, classification, and generation settings spanning diverse multimodal tasks. All experiments use grouped parameter-efficient adaptation with instruction-conditioned gating.
Additional implementation details are provided in Appendix~\ref{app:implementation_details}.

\subsection{Tasks and Datasets}
We evaluate GUIDE across multimodal reasoning, classification, and generation benchmarks selected for their structured and perturbable evidence categories.

\noindent\textbf{Multimodal Reasoning.}
We evaluate on GQA~\cite{hudson2019gqa} and TextVQA~\cite{singh2019towards}.
GQA enables analysis of semantic, relational, appearance, and contextual evidence under controlled visual reasoning settings, while TextVQA provides a complementary setting requiring explicit OCR and textual grounding signals.
For GQA, we evaluate on the balanced split to reduce shortcut-associated answer biases.

\noindent\textbf{Multimodal Classification.}
We evaluate on MM-IMDb~\cite{arevalo2017gatedmultimodalunitsinformation}, CREMA-D~\cite{cremad}, and RAVDESS~\cite{ravdess}.
These datasets contain diverse evidence sources and shortcut-associated cues spanning appearance, context, motion, acoustic, and demographic signals, making them suitable for studying controllable evidence reliance under fixed task semantics.

\noindent\textbf{Multimodal Generation.}
We evaluate on Flickr30K~\cite{flickr30k} image captioning.
Caption generation provides a complementary setting in which evidence modulation must persist throughout autoregressive decoding rather than affecting only a single prediction step.
We follow standard benchmark splits for all datasets and perform perturbation-based evaluation only on held-out evaluation partitions.

\subsection{Instruction Taxonomy}
\vspace{-0.3em}
We construct instruction families corresponding to different evidence preferences, including semantic-focused, appearance-focused, background-suppression, motion-focused, and OCR-focused conditions depending on the benchmark.
In addition to fixed templates, we evaluate paraphrased and compositional instruction variants to test generalization beyond fixed lexical formulations.
Representative instruction templates are provided in Appendix~\ref{app:instruction_taxonomy}.

\subsection{Evidence Perturbation Protocol}
\vspace{-0.3em}
To measure functional reliance on different evidence sources, we use taxonomy-aligned perturbation protocols targeting semantic, appearance, contextual, OCR/textual, motion, and acoustic evidence categories depending on the benchmark.
These interventions include modality-specific masking, corruption, replacement, and suppression operations designed to selectively degrade targeted evidence while preserving overall task solvability.
For perturbations requiring coarse evidence localization, we use pretrained multimodal backbones to obtain weak evidence-region proposals without additional supervision or task-specific fine-tuning.
Additional perturbation details and backbone configurations are provided in Appendix~\ref{app:backbones}.

\subsection{Evaluation Metrics}
\vspace{-0.3em}
We evaluate GUIDE using both task-level performance metrics and pathway-level evidence-modulation metrics.
To quantify instruction-conditioned evidence modulation, we report:
(i) Gate Modulation Response (GMR), which measures changes in pathway activations between instructions, and
(ii) Reliance Sensitivity (RS), which measures functional dependence on targeted evidence sources under controlled perturbations.
GMR and RS provide complementary measures of pathway activation and functional evidence reliance.
We additionally report standard task-specific metrics for each benchmark to evaluate task preservation under instruction-conditioned evidence modulation.

\vspace{-0.3em}
\section{Results}
\begin{figure}[t]
   \centering
   \includegraphics[width=\columnwidth]{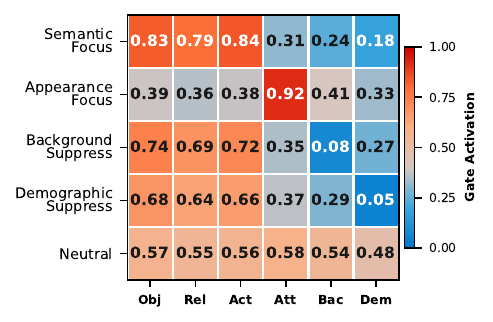}
   \vspace{-2em}
   \caption{Gate activation heatmap on GQA under different instruction conditions. Obj, Rel, Act, Att, Bac, Dem denote \textbf{Obj}ect, \textbf{Rel}ation, \textbf{Act}ion, \textbf{Att}ribute, \textbf{Bac}kground, \textbf{Dem}ographic respectively}
   \label{fig:heatmap_gqa}
   
\end{figure}
\vspace{-0.5em}
\subsection{Controlled Evidence Routing}
\vspace{-0.5em}
We first evaluate whether instruction-conditioned gating produces structured modulation of evidence pathways.
Figure~\ref{fig:heatmap_gqa} shows gate activation patterns on GQA under different instruction conditions, while Figure~\ref{fig:heatmap_mmimdb}--Figure~\ref{fig:heatmap_flickr} present corresponding results on additional multimodal benchmarks.

Across all datasets, GUIDE produces consistent and instruction-aligned changes in pathway activations.
Semantic-focused instructions increase activation of semantic and relational pathways, whereas appearance-focused instructions increase activation of appearance-related pathways.
Similarly, suppression-oriented instructions selectively reduce activation of the targeted pathway while preserving task-relevant pathways.
Neutral instructions produce comparatively balanced activation patterns, indicating less pathway-specific modulation in the absence of an explicit evidence preference.
These instruction-conditioned patterns remain consistent across multimodal reasoning, classification, and generation settings despite differences in modality and task structure.
Overall, GUIDE induces structured and instruction-consistent modulation of evidence pathways rather than only superficial prompt-conditioned output variation.

\subsection{Functional Evidence Redistribution}
\begin{figure}[t]
   \centering
   \includegraphics[width=\columnwidth]{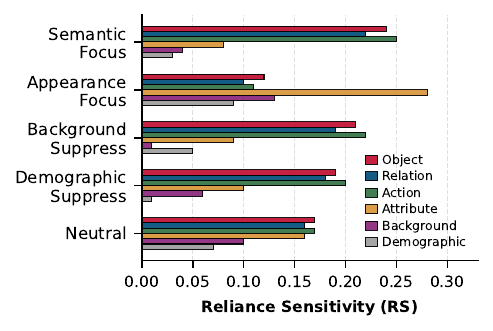}
   \vspace{-2.2em}
   \caption{RS under different instruction conditions on GQA.
Instructions selectively redistribute functional dependence across evidence pathways.}
   \label{fig:grouped_barplot_gqa}
\end{figure}

We next evaluate whether instruction-conditioned pathway modulation is accompanied by corresponding changes in functional evidence reliance.
Figure~\ref{fig:grouped_barplot_gqa} reports Reliance Sensitivity (RS) across evidence pathways on GQA under different instruction conditions, while Figure~\ref{fig:grouped_barplot_mmimdb}--Figure~\ref{fig:grouped_barplot_flickr} present corresponding results on additional benchmarks.
Across tasks, RS patterns consistently align with the evidence preferences specified by the instructions.
Semantic-focused instructions increase prediction sensitivity to semantic, relational, and interaction-related evidence, whereas appearance-focused instructions increase sensitivity to appearance-related evidence.
Suppression-oriented instructions reduce sensitivity to the targeted evidence source while largely preserving dependence on task-relevant pathways.
For example, background-suppression instructions consistently reduce functional dependence on contextual evidence without substantially reducing overall prediction sensitivity.
These results show that instruction-conditioned pathway modulation is accompanied by systematic changes in the evidence sources on which model predictions functionally depend.

\subsection{Counterfactual Evidence Control}
We next evaluate whether instruction-conditioned evidence modulation is associated with corresponding changes in robustness under controlled perturbations.
Figure~\ref{fig:gqa_examples} and \ref{fig:textvqa_examples} provide qualitative examples showing that different instructions induce distinct evidence emphasis patterns while preserving the same final prediction.
Under semantic-focused instructions, the model places greater emphasis on action-relevant and semantically meaningful regions, whereas appearance-focused instructions emphasize figure-centric visual attributes.

\begin{figure}[t]
   \centering
   \includegraphics[width=\columnwidth]{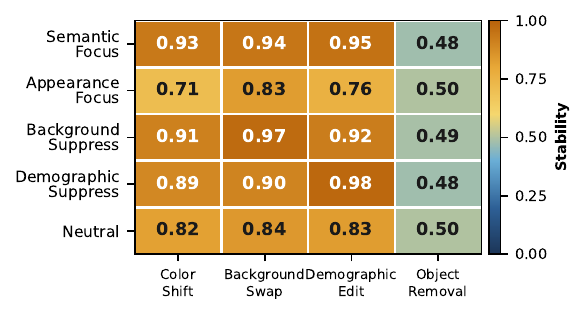}
   \caption{Counterfactual stability under different instruction conditions and perturbation types on GQA.
Semantic-focused and background-suppression instructions remain robust under appearance and contextual perturbations, whereas object removal substantially reduces stability across all settings.}
   \label{fig:counterfactual_stability_gqa}
      \vspace{-1.5em}
\end{figure}

To quantify this behavior, Figure~\ref{fig:counterfactual_stability_gqa} reports counterfactual stability under four perturbation types.
Semantic-focused and background-suppression instructions remain highly stable under color and background perturbations, whereas appearance-focused instructions exhibit substantially lower robustness under appearance-related shifts.
In contrast, object removal consistently reduces stability across all instruction conditions, indicating that object-level evidence remains task-critical across routing preferences.

\begin{figure}[t]
   \centering
   \includegraphics[width=\columnwidth]{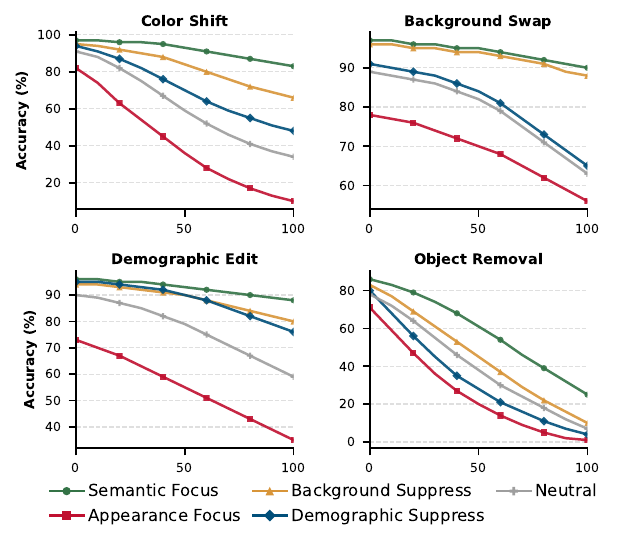}
   \caption{Prediction robustness under increasing perturbation strength on GQA.
Appearance-focused instructions degrade most rapidly under color shifts, whereas semantic-focused and background-suppression instructions remain comparatively robust.}
\vspace{-1em}
   \label{fig:sensitivity_curve_row2}
\end{figure}

Figure~\ref{fig:sensitivity_curve_row2} further analyzes robustness under progressively increasing perturbation strength.
Appearance-focused instructions exhibit the steepest degradation under color shifts, while semantic-focused and background-suppression instructions retain substantially higher accuracy throughout the perturbation range.
Additional perturbation stability analyses across MM-IMDb and Flickr30K are provided in Appendix~\ref{app:perturbation_stability}.
All robustness curves are averaged across the evaluation set over multiple perturbation strengths.
Together, these results show that instruction-conditioned pathway modulation is accompanied by systematic changes in functional dependence on targeted evidence sources.

\begin{figure}[!t]
   \centering
   \includegraphics[width=\columnwidth]{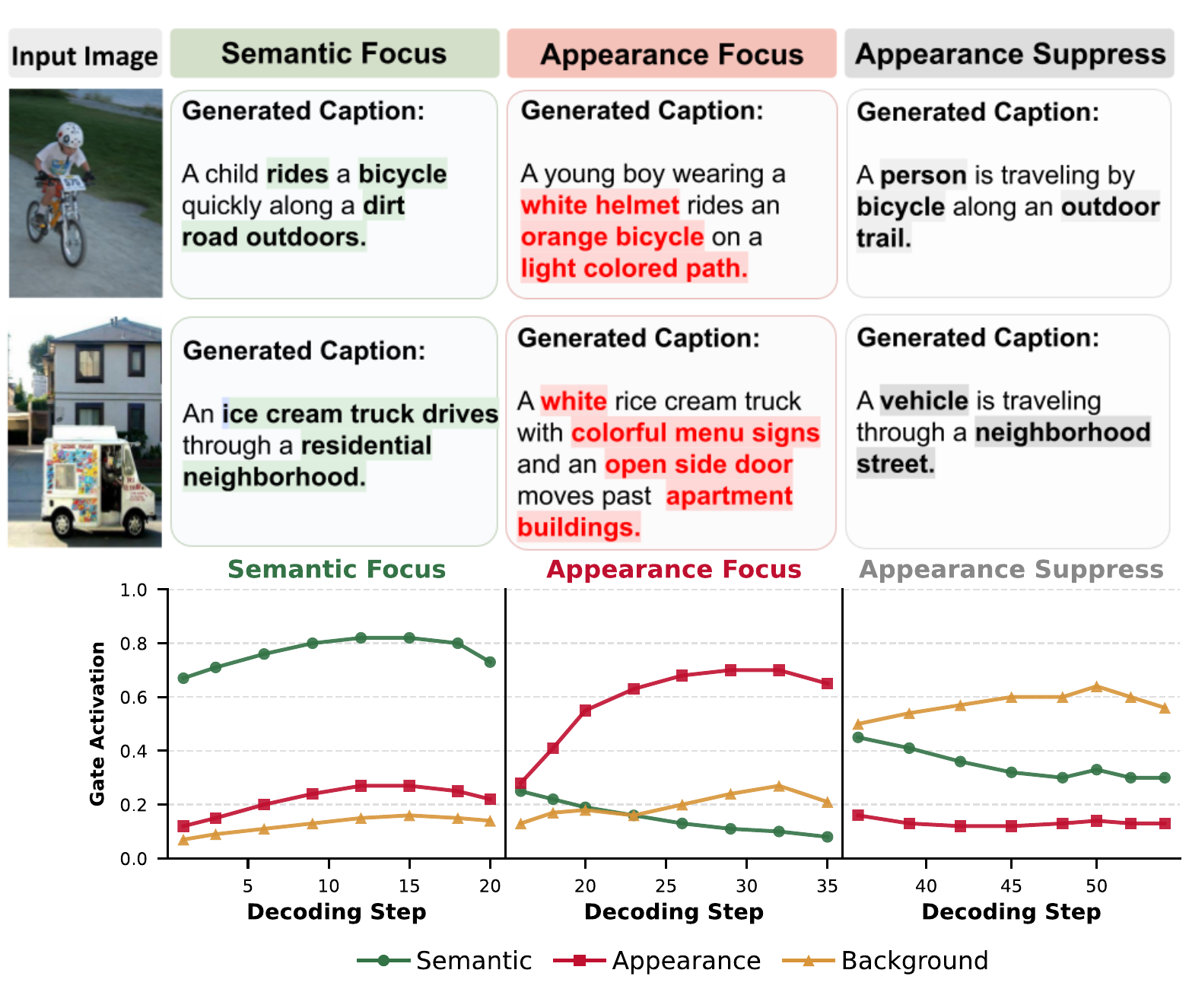}
   \caption{Instruction-conditioned decoding-time evidence routing on Flickr30K. Different instructions induce persistent changes in pathway activations throughout autoregressive generation, resulting in captions emphasizing either semantic/appearance attributes.}
\vspace{-1.0em}
   \label{fig:caption_gate_dynamic}
\end{figure}

\vspace{-0.5em}
\subsection{Decoding-Time Evidence Routing} Figure~\ref{fig:caption_gate_dynamic} illustrates pathway-level gate activations across autoregressive decoding steps on Flickr30K under different instruction conditions.
Distinct instructions produce consistently different pathway activation dynamics throughout generation rather than only at initialization.
Under semantic-focused instructions, semantic pathways remain strongly activated across decoding steps, resulting in captions that prioritize actions and scene interactions.
In contrast, appearance-focused instructions increase activation of appearance-related pathways, leading to captions that emphasize visual attributes such as clothing, color, and object appearance.
Appearance-suppression instructions selectively reduce appearance-related gate activation while preserving semantic and contextual pathways, resulting in captions that retain scene-level semantics while providing fewer appearance-specific details.

Importantly, these instruction-conditioned activation patterns persist throughout autoregressive decoding, indicating that GUIDE modulates pathway contributions during generation rather than only biasing the initial decoding state.
The plotted gate trajectories correspond to representative decoding sequences and show similar trends across additional evaluation examples.
Additional analyses of compositional and conflict-aware routing are provided in Appendix~\ref{app:compositional_routing}.

\subsection{Same Answer, Different Evidence}
We evaluate whether GUIDE can preserve task behavior while substantially altering internal evidence reliance.
Table~\ref{tab:same_answer_diff_evidence} reports output consistency together with changes in pathway modulation (GMR) and changes in functional reliance ($\Delta$RS) across conflicting instruction conditions.
Across all benchmarks, GUIDE maintains high output consistency despite substantial changes in pathway activation and functional reliance.
For example, GQA preserves the same answer in 93.8\% of cases while producing substantial shifts in both pathway modulation and functional reliance.
These results provide direct quantitative evidence that GUIDE can alter how multimodal models internally use evidence without necessarily changing the final prediction. Additional analyses are provided in Appendix~\ref{app:conflict_tables} and ~\ref{app:layerwise}.

\begin{table}[!t]
\centering	
\small	
\caption{
Same-answer behavior under semantic--appearance evidence-routing shifts.
GUIDE preserves high output consistency despite substantial pathway modulation (GMR) and changes in functional reliance ($\Delta$RS).
Flickr30K reports caption similarity (0--1), while other datasets report same-prediction rate (\%).
}	
\label{tab:same_answer_diff_evidence}
\begin{tabular}{lccc}	
\toprule	
\textbf{Dataset} & \textbf{Output Consistency} & \textbf{GMR} & \textbf{$\Delta$RS} \\	
\midrule	
GQA & 93.8 & 0.61 & 0.19 \\	
TextVQA & 91.8 & 0.51 & 0.16 \\	
MM-IMDb & 92.1 & 0.58 & 0.17 \\	
CREMA-D & 91.7 & 0.55 & 0.18 \\	
RAVDESS & 91.1 & 0.53 & 0.17 \\	
Flickr30K & 0.84 & 0.63 & 0.20 \\	
\bottomrule	
\end{tabular}	
\vspace{-1em}
\end{table}

\subsection{Ablation Study}
\vspace{-0.3em}
We evaluate the contribution of grouped routing and dynamic instruction-conditioned gating.
Table~\ref{tab:ablation} reports representative ablation results on GQA, CREMA-D, and Flickr30K.
Removing taxonomy-aligned grouped pathways substantially reduces both Gate Modulation Response (GMR) and changes in functional reliance ($\Delta$RS), indicating that structured pathway decomposition plays an important role in controllable evidence modulation.
Replacing dynamic instruction-conditioned gates with static routing also consistently decreases pathway modulation, functional reliance shifts, and perturbation robustness.
Prompt-only control exhibits the weakest evidence-modulation behavior across all three settings.
Overall, these results show that grouped adaptation pathways and dynamic instruction-conditioned gating jointly contribute to effective multimodal evidence control.
Full ablation results across all benchmarks are provided in Appendix~\ref{app:full_ablation}.

\begin{table}[t]
\centering
\small
\setlength{\tabcolsep}{5pt}
\renewcommand{\arraystretch}{0.95}
\caption{
Representative ablation results across multimodal reasoning (GQA), classification (CREMA-D), and generation (Flickr30K).
Removing grouped pathways or replacing dynamic instruction-conditioned gates reduces pathway modulation (GMR), changes in functional reliance ($\Delta$RS), and perturbation robustness (Robust.).
Full benchmark results are provided in Appendix~\ref{app:full_ablation}.
}
\label{tab:ablation}

\begin{tabular}{llccc}
\toprule
\textbf{Dataset} & \textbf{Variant} & \textbf{GMR}$\uparrow$ & \textbf{$\Delta$RS}$\uparrow$ & \textbf{Robust.}$\uparrow$ \\
\midrule

\multirow{4}{*}{GQA}
& Full & \textbf{0.61} & \textbf{0.19} & \textbf{0.93} \\
& No grouped & 0.22 & 0.07 & 0.81 \\
& Static gates & 0.38 & 0.10 & 0.86 \\
& Prompt-only & 0.18 & 0.06 & 0.79 \\

\midrule

\multirow{4}{*}{CREMA-D}
& Full & \textbf{0.55} & \textbf{0.18} & \textbf{0.91} \\
& No grouped & 0.19 & 0.06 & 0.79 \\
& Static gates & 0.33 & 0.09 & 0.84 \\
& Prompt-only & 0.16 & 0.05 & 0.77 \\

\midrule

\multirow{4}{*}{Flickr30K}
& Full & \textbf{0.63} & \textbf{0.20} & \textbf{0.84} \\
& No grouped & 0.23 & 0.07 & 0.73 \\
& Static gates & 0.39 & 0.11 & 0.78 \\
& Prompt-only & 0.19 & 0.06 & 0.71 \\

\bottomrule
\end{tabular}
\vspace{-0.5em}
\end{table}

\vspace{-0.8em}
\section{Conclusion}
\vspace{-0.5em}
We introduced GUIDE, a framework for controlling multimodal evidence usage through natural language instructions.
By combining taxonomy-aligned grouped adaptation pathways with instruction-conditioned gating, GUIDE enables models to modulate different forms of evidence during reasoning and generation while largely preserving task behavior.
Across multimodal reasoning, classification, and generation benchmarks, GUIDE produces consistent instruction-aligned pathway modulation, corresponding shifts in functional evidence reliance, and improved robustness under targeted evidence perturbations.
We further show that similar outputs can be maintained under substantially different evidence-reliance patterns, and that these instruction-conditioned effects persist throughout autoregressive decoding.
Together, these findings suggest that multimodal instruction following can extend beyond controlling what models output toward regulating how different evidence sources contribute to model predictions.
\begin{figure*}[p]
   \centering
   \includegraphics[width=\textwidth]{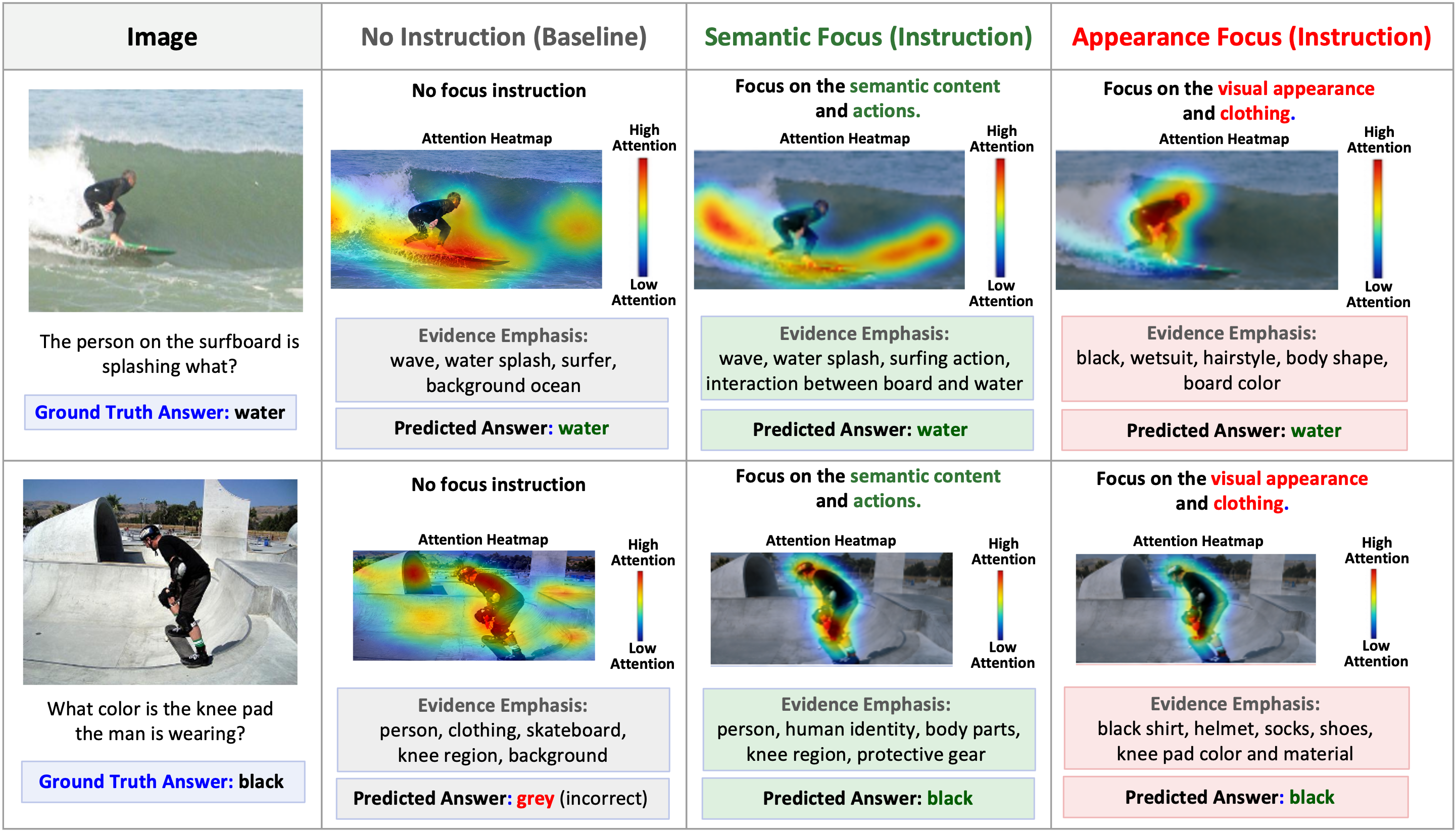}
   \caption{Qualitative examples of instruction-conditioned evidence routing on GQA.
Semantic-focused instructions emphasize action-relevant and interaction-based evidence, whereas appearance-focused instructions prioritize visual attribute cues while preserving the same final prediction.}
   \label{fig:gqa_examples}
\end{figure*}
\begin{figure*}[p]
   \centering
   \includegraphics[width=\textwidth]{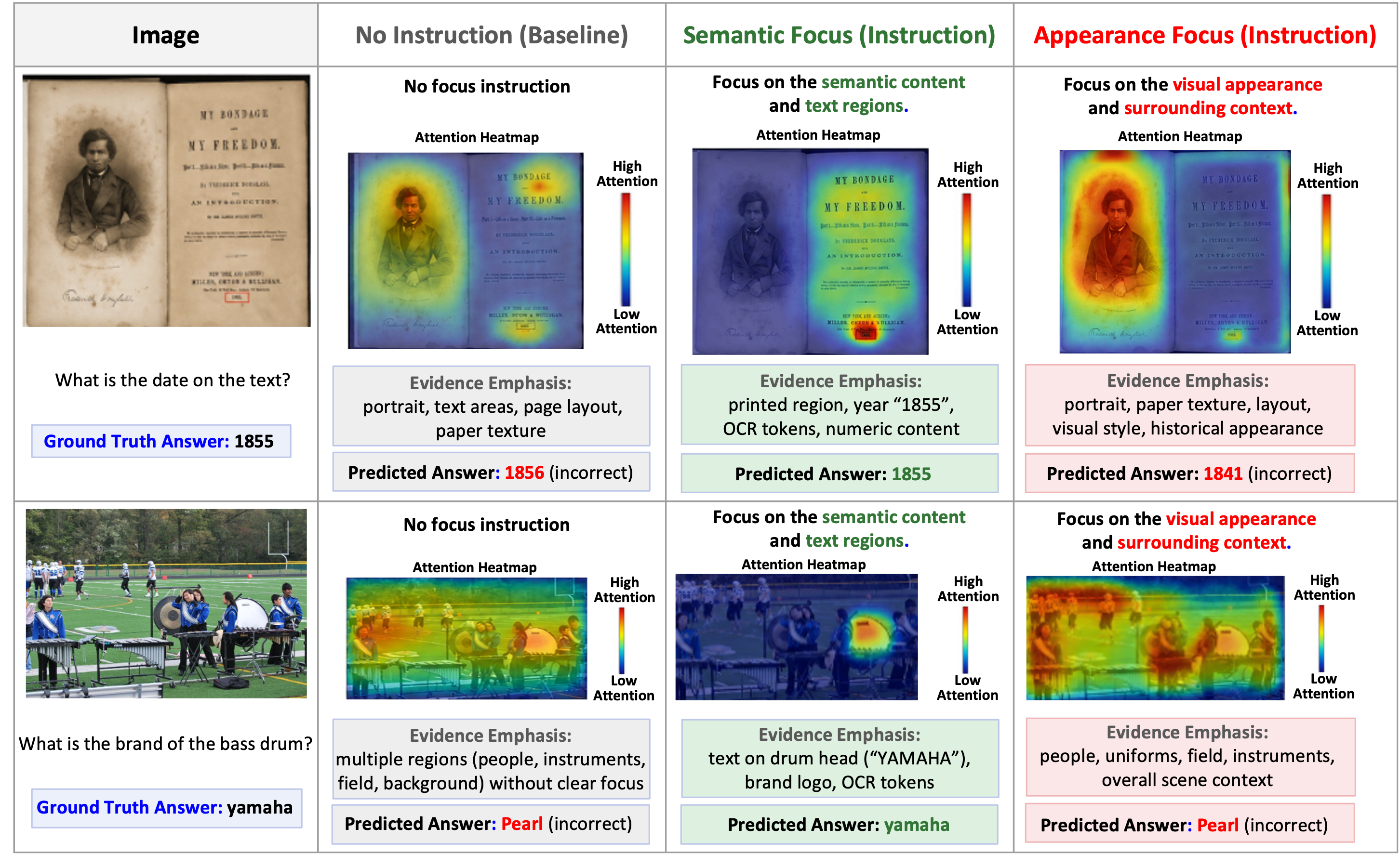}
   \caption{
Qualitative examples of instruction-conditioned evidence routing on TextVQA.
Semantic-focused instructions increase reliance on OCR and text-relevant regions, whereas appearance-focused instructions emphasize visual appearance and contextual cues, often leading to different prediction behavior under the same input.
}
   \label{fig:textvqa_examples}
\end{figure*}

\section*{Limitations}
GUIDE does not explicitly disentangle semantic factors or guarantee perfectly isolated evidence pathways.
Instead, the proposed grouped routing mechanism provides an operational intervention space for modulating functional evidence reliance under natural language instructions.
As a result, pathway activations should not be interpreted as uniquely corresponding to fully disentangled semantic concepts.
In addition, the evidence perturbation protocols used in this work are approximate operational interventions rather than precise semantic manipulations.
Although we employ controlled perturbations and weak localization strategies, some perturbations may partially affect multiple evidence categories simultaneously.
Future work may explore more principled causal intervention and localization methods for multimodal evidence control.
Finally, GUIDE is evaluated primarily on controlled multimodal reasoning, classification, and generation benchmarks.
Additional evaluation on larger-scale open-ended multimodal agents and real-world deployment settings remains an important direction for future work.

\section*{Potential Risks}
Techniques for controllable evidence routing may potentially be misused to selectively amplify or suppress specific evidence sources in ways that obscure model reasoning behavior.
For example, evidence suppression mechanisms could unintentionally hide undesirable biases while preserving similar output behavior.
In addition, imperfect evidence localization and perturbation protocols may introduce unintended artifacts or incomplete suppression of sensitive cues.
We therefore emphasize that GUIDE should not be interpreted as providing guaranteed fairness, bias removal, or fully interpretable reasoning.
Instead, the framework is intended as a research tool for studying controllable multimodal evidence reliance under operational intervention settings.

\section*{Ethics Statement}
Human evaluation was conducted using voluntary annotators recruited for research purposes.
Annotators were informed about the evaluation procedure and provided consent prior to participation.
The annotation tasks involved evaluating multimodal model outputs and did not involve sensitive personal data, medical information, or high-risk decision making.
All data used in human evaluation were obtained from publicly available research benchmarks under their original licenses.
No personally identifiable information was collected or stored during the annotation process.

\section*{Acknowledgement}
We thank Prof. Eduard Hovy and Prof. Sangwook Yi for their support of this research. This research was supported by the Institute of Information \& communications Technology Planning \& Evaluation (IITP) grant funded by the Korea government (MSIT) (No.RS-2026-25519206, Development of human-like intelligent generative agents based on interactive multimodal reverse prompting), and IITP grant funded by the Korea government (MSIT) (No.RS-2025-02217259, Development of self-evolving AI bias detection-correction-explain platform based on international multidisciplinary governance).

\bibliography {custom}


\newpage
\appendix

\clearpage
\section{Dataset-Specific Evidence Taxonomy}
\label{app:taxonomy_construction}
GUIDE uses dataset-specific evidence taxonomies to define controllable multimodal evidence pathways aligned with the dominant evidence structure of each benchmark.
Rather than assuming perfectly disentangled semantic factors, the taxonomy provides coarse functional partitions corresponding to task-relevant multimodal cues that may contribute differently to model predictions.
The taxonomy is constructed based on three principles:
(1) identifying primary evidence sources relevant to successful task completion,
(2) separating shortcut-associated or potentially confounding evidence categories,
and
(3) defining evidence categories that can be targeted through controlled perturbations.
The taxonomy distinguishes between \emph{high-level evidence families} and \emph{operational routing groups}.
A high-level family may be instantiated by multiple fine-grained routing groups depending on the benchmark.
For example, in GQA, semantic evidence is decomposed into Object, Relation, and Action groups, while appearance-related evidence is represented by an Attribute group, together with Background and Demographic groups.
This hierarchical organization allows GUIDE to retain a common evidence taxonomy while adapting the operational routing granularity to each task.

Table~\ref{tab:perturbation_protocols} summarizes the perturbation protocols associated with each evidence category.
These perturbations are designed to measure relative changes in functional reliance under targeted evidence degradation rather than to simulate naturally occurring corruptions.
For visual reasoning benchmarks such as GQA and Flickr30K, the high-level taxonomy includes semantic interaction cues, appearance attributes, environmental context, and person-related or demographic appearance signals.
Semantic evidence primarily corresponds to object interactions, action understanding, and relational reasoning, while appearance evidence captures visual attributes such as clothing color, texture, and style.
Background evidence captures environmental and scene-level contextual information that may support prediction.
Person-related or demographic evidence captures facial and identity-associated appearance cues that may correlate spuriously with task outputs.
For TextVQA, the taxonomy additionally includes an OCR/Text family corresponding to explicit textual grounding signals.
This provides a distinct evidence source from object appearance or scene context and enables targeted analysis of textual versus visual reliance.
For MM-IMDb, the taxonomy includes semantic narrative information, interaction-related character relationships, poster appearance cues, contextual visual style, and demographic-related appearance signals.
This formulation enables controlled analysis of whether models rely on narrative semantics or shortcut-associated visual correlations contained in movie posters.
For audiovisual emotion benchmarks such as CREMA-D and RAVDESS, the taxonomy includes facial-expression signals, acoustic emotion cues, motion dynamics, appearance information, and contextual background signals.
This organization enables analysis of how language instructions modulate reliance across visual, acoustic, and temporal evidence while preserving emotion classification behavior.
Importantly, these evidence categories are not treated as perfectly independent causal mechanisms.
Instead, they provide structured operational partitions for pathway modulation and perturbation-based analysis of functional evidence reliance across diverse multimodal settings.

\begin{table*}[t]
\centering
\small
\caption{
Dataset-specific perturbation protocols used for evaluating functional reliance on different evidence pathways.
Perturbations are designed as controlled interventions targeting task-relevant evidence categories. Perturbations are applied only during evaluation and are not used for training GUIDE.
}
\label{tab:perturbation_protocols}
\setlength{\tabcolsep}{4pt}
\renewcommand{\arraystretch}{1.0}
\begin{tabular}{lll}
\toprule
\textbf{Dataset} & \textbf{Evidence Category} & \textbf{Perturbation Protocol} \\
\midrule

\multirow{4}{*}{GQA}
& Semantic 
& Object masking, relation disruption, interaction-region masking, action-region occlusion \\

& Appearance 
& Color jitter, texture perturbation, attribute-preserving recoloring \\

& Background 
& Contextual background replacement, environmental region masking, scene blurring \\

& Demographic 
& Facial anonymization, skin-tone perturbation, hairstyle masking \\

\midrule

\multirow{4}{*}{TextVQA}
& Semantic 
& Object-region masking and interaction-region perturbation \\

& OCR/Text 
& OCR-region masking, character corruption, localized text blurring \\

& Appearance 
& Color perturbation and texture modification \\

& Background 
& Contextual scene replacement and environmental masking \\

\midrule

\multirow{5}{*}{MM-IMDb}
& Semantic 
& Plot-summary truncation, semantic phrase removal, narrative corruption \\

& Interaction 
& Character-relation sentence removal and interaction phrase masking \\

& Appearance 
& Poster recoloring, texture modification, visual style perturbation \\

& Background 
& Poster background replacement and contextual style masking \\

& Demographic 
& Face-region perturbation and identity-related appearance masking \\

\midrule

\multirow{5}{*}{CREMA-D}
& Facial Expression 
& Eye/mouth-region masking and expression suppression \\

& Vocal Emotion 
& Pitch perturbation, spectral masking, temporal audio corruption \\

& Motion 
& Temporal frame shuffle, motion blur, frame corruption \\

& Appearance 
& Clothing/color perturbation and illumination modification \\

& Background 
& Background scene masking and environmental audio suppression \\

\midrule

\multirow{5}{*}{RAVDESS}
& Facial Emotion 
& Facial-region masking and expression suppression \\

& Acoustic Tone 
& Pitch shifting, spectral masking, temporal audio perturbation \\

& Motion 
& Motion blur and temporal frame corruption \\

& Appearance 
& Clothing/style perturbation and illumination modification \\

& Context 
& Background suppression and contextual scene masking \\

\midrule

\multirow{5}{*}{Flickr30K}
& Semantic 
& Foreground object masking and semantic object removal \\

& Interaction 
& Contact-region masking and interaction-region occlusion \\

& Appearance 
& Clothing recoloring, texture perturbation, color jitter \\

& Background 
& Scene-context replacement while preserving foreground entities \\

& Demographic 
& Facial anonymization and identity-cue suppression \\

\bottomrule
\end{tabular}
\end{table*}

\section{Instruction Taxonomy and Pathway Conditioning}
\label{app:instruction_taxonomy}
\vspace{-0.4em}

GUIDE uses dataset-specific instruction templates to provide semantically consistent conditioning signals for different evidence preferences across multimodal benchmarks.
The templates are designed to encourage emphasis or suppression of task-relevant evidence categories rather than optimize benchmark-specific prompt engineering behavior.
Table~\ref{tab:instruction_templates} shows representative instruction templates used during training and evaluation.
Multiple paraphrased variants were constructed for each evidence preference to reduce sensitivity to fixed prompt wording.
Base instruction templates were used during both training and evaluation, while paraphrased variants were additionally evaluated to assess generalization beyond fixed lexical formulations.

\begin{table*}[t]
\centering
\small
\caption{Representative instruction templates used for dataset-specific evidence-pathway conditioning.}
\label{tab:instruction_templates}
\setlength{\tabcolsep}{4pt}
\renewcommand{\arraystretch}{1.0}
\begin{tabular}{llll}
\toprule
\textbf{Dataset} & \textbf{Pathway} & \textbf{Instruction Type} & \textbf{Example Instruction} \\
\midrule

\multirow{4}{*}{GQA}
& Semantic 
& Focus 
& ``Focus on actions and object interactions.'' \\

& Appearance 
& Focus 
& ``Focus on visual appearance and clothing attributes.'' \\

& Appearance 
& Suppress 
& ``Ignore clothing and appearance-related cues.'' \\

& Background 
& Suppress 
& ``Ignore environmental context and focus on foreground interactions.'' \\

\midrule

\multirow{3}{*}{TextVQA}
& OCR/Text 
& Focus 
& ``Rely primarily on textual evidence in the image.'' \\

& OCR/Text 
& Suppress 
& ``Ignore textual content and rely on visual context.'' \\

& Appearance 
& Suppress 
& ``Ignore visual appearance and focus on readable text.'' \\

\midrule

\multirow{3}{*}{MM-IMDb}
& Semantic 
& Focus 
& ``Focus on narrative and plot-related information.'' \\

& Interaction 
& Focus 
& ``Focus on character relationships and interactions.'' \\

& Appearance 
& Suppress 
& ``Ignore poster style and rely on semantic content.'' \\

\midrule

\multirow{3}{*}{CREMA-D}
& Vocal Emotion 
& Focus 
& ``Focus on vocal tone and acoustic emotion cues.'' \\

& Facial Expression 
& Suppress 
& ``Ignore facial appearance and rely on speech emotion.'' \\

& Motion 
& Focus 
& ``Focus on motion dynamics and temporal behavior.'' \\

\midrule

\multirow{3}{*}{RAVDESS}
& Acoustic Tone 
& Focus 
& ``Focus on vocal tone and acoustic emotion cues.'' \\

& Motion 
& Focus 
& ``Focus on temporal motion and behavioral dynamics.'' \\

& Appearance 
& Suppress 
& ``Ignore clothing and appearance-related cues.'' \\

\midrule

\multirow{3}{*}{Flickr30K}
& Interaction 
& Focus 
& ``Focus on human-object interactions.'' \\

& Appearance 
& Suppress 
& ``Ignore clothing and appearance-related attributes.'' \\

& Background 
& Suppress 
& ``Ignore background context and focus on foreground entities.'' \\

\bottomrule
\end{tabular}
\end{table*}

\section{GMR Computation Details}
\label{app:gmr_details}
\vspace{-0.4em}

GMR measures the magnitude of instruction-conditioned modulation across taxonomy-aligned routing groups.
For each input sample, we compute gate activations under two contrasting instructions applied to the same multimodal input.
Typical instruction pairs include semantic-focused versus appearance-focused prompts, OCR-focused versus contextual prompts, or motion-focused versus appearance-focused prompts, depending on the benchmark.

For autoregressive generation tasks, gate activations are recorded at each decoding step and across gated transformer layers.
Pathway-wise activation differences between the two instruction conditions are first averaged over decoding steps within each layer and then aggregated across gated layers.
The resulting pathway differences are averaged over the $C$ operational routing groups, consistent with the GMR definition in Section~\ref{sec:guide}.
For classification tasks, gate activations are extracted from the final prediction representation and aggregated across gated transformer layers; no temporal decoding aggregation is required.

All reported GMR values are finally averaged across the evaluation set to reduce variation from individual samples or instruction instances.
Higher GMR indicates stronger instruction-conditioned changes in pathway activation patterns.

\vspace{-0.5em}
\section{Training Objective}
\label{app:training_objective}

GUIDE is trained using the original task objective while learning instruction-conditioned evidence modulation through grouped adaptation pathways and dynamic gating.
Given input $x$, instruction $I$, and target output $y$, the model is optimized using the standard task loss:
\[
\mathcal{L}_{\text{task}}
=
-\log p(y \mid x, I).
\]

During training, only the grouped low-rank adaptation parameters and gating networks are updated, while the pretrained backbone remains frozen.
This preserves the pretrained multimodal capabilities while enabling lightweight instruction-conditioned modulation through the added adaptation pathways.

The gating network is trained jointly with the task objective without explicit supervision over pathway activations.
Instead, GUIDE learns instruction-sensitive pathway modulation through task optimization across taxonomy-aligned grouped adaptation pathways, without requiring manually annotated evidence labels or explicitly disentangled internal representations.

Because GUIDE uses independent sigmoid gates, each pathway activation $g_c \in [0,1]$ is treated independently rather than being normalized across pathways.
To encourage confident and selective pathway activation, we apply a Bernoulli entropy regularization term:
\[
\mathcal{L}_{\text{gate}}
=
-\lambda_{\mathrm{gate}} 
\sum_{c=1}^{C}
\left[
g_c \log g_c
+
(1-g_c)\log(1-g_c)
\right],
\]
where $\lambda_{\mathrm{gate}}$ controls the strength of gate regularization.
The final training objective is:
\[
\mathcal{L}
=
\mathcal{L}_{\text{task}}
+
\mathcal{L}_{\text{gate}}.
\]

This objective preserves the original task supervision while encouraging selective instruction-conditioned modulation of taxonomy-aligned routing groups.

\section{Extended Conflict Routing Results}
\label{app:conflict_tables}

To further evaluate whether GUIDE supports compositional evidence control beyond fixed instruction pairs, we analyze additional contrasting evidence-routing conditions across multimodal reasoning, classification, and generation tasks.
Table~\ref{tab:same_answer_diff_evidence_full} reports output consistency together with pathway modulation (GMR) and changes in functional reliance ($\Delta$RS) across multiple instruction pairs.

Across datasets, GUIDE consistently preserves high output consistency despite substantial changes in pathway activation and functional reliance.
Different instruction combinations induce distinct magnitudes of pathway modulation depending on the targeted evidence categories.
Contrasts between semantic- and appearance-focused instructions generally produce the largest modulation, whereas neutral or partially aligned instruction pairs yield comparatively smaller changes.

Importantly, GUIDE maintains relatively stable task behavior even under strongly contrasting evidence preferences across visual, textual, acoustic, and generation settings.
These results further support that GUIDE enables continuous and compositional evidence modulation rather than relying on isolated prompt templates or single-pathway switching behavior.

\section{Instruction Conflict and Compositional Evidence Routing}
\label{app:compositional_routing}
\begin{figure}[h]
   \centering
   \includegraphics[width=\columnwidth]{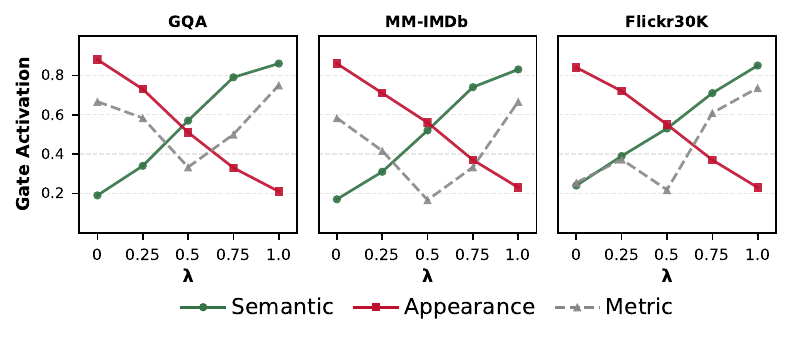}
   \caption{Instruction-conditioned semantic--appearance trade-off curves across GQA, MM-IMDb, and Flickr30K.
   As instruction emphasis shifts toward semantic reasoning, GUIDE smoothly modulates semantic and appearance pathway activations while maintaining relatively stable task performance.
   Dashed gray curves denote task consistency metrics for GQA and MM-IMDb, and CIDEr scores for Flickr30K.}
   \vspace{-0.5em}
   \label{fig:conflict_routing}
\end{figure}

We next evaluate whether GUIDE can accommodate contrasting evidence preferences within a single instruction.
To this end, we construct compositional instructions such as \textit{``focus on semantic relationships while ignoring appearance''} and \textit{``focus on appearance while suppressing semantic interactions.''}
Figure~\ref{fig:conflict_routing} shows semantic--appearance trade-off behavior on GQA, MM-IMDb, and Flickr30K, where $\alpha$ controls the relative emphasis
between semantic-focused and appearance-focused evidence preferences.
As $\alpha$ shifts from appearance-oriented to semantic-oriented reasoning, GUIDE produces smooth changes in semantic and appearance pathway activations rather than abrupt switching behavior.
Balanced conditions yield intermediate activation patterns across the two pathway types, indicating compositional evidence modulation rather than exclusive single-pathway selection.
Despite substantial changes in pathway activation, GUIDE maintains relatively stable task behavior, supporting continuous and conflict-aware evidence modulation beyond fixed instruction templates.

\begin{table*}[!t]
    \centering
    \small
    \caption{
    Same-answer behavior under contrasting evidence-routing instructions across multimodal tasks.
    GUIDE preserves high output consistency despite substantial pathway modulation (GMR) and changes in functional reliance ($\Delta$RS).
    Flickr30K reports caption similarity (0--1), while other datasets report same-prediction rate (\%).
    }
    \label{tab:same_answer_diff_evidence_full}
    \begin{tabular}{lcccc}
    \toprule
    \textbf{Dataset} & \textbf{Instruction Pair} & \textbf{Output Consistency} & \textbf{GMR} & \textbf{$\Delta$RS} \\
    \midrule
    \multirow{5}{*}{GQA}
    & Semantic $\leftrightarrow$ Appearance & 93.8 & 0.61 & 0.19 \\
    & Semantic $\leftrightarrow$ Background Suppress & 95.1 & 0.47 & 0.14 \\
    & Appearance $\leftrightarrow$ Background Suppress & 91.4 & 0.54 & 0.17 \\
    & Semantic $\leftrightarrow$ Demographic Suppress & 94.6 & 0.42 & 0.12 \\
    & Neutral $\leftrightarrow$ Semantic & 89.7 & 0.31 & 0.09 \\
    \midrule
    \multirow{5}{*}{TextVQA}
    & Semantic $\leftrightarrow$ OCR/Text & 92.6 & 0.57 & 0.18 \\
    & Semantic $\leftrightarrow$ Appearance & 91.8 & 0.51 & 0.16 \\
    & OCR/Text $\leftrightarrow$ Background Suppress & 93.4 & 0.49 & 0.15 \\
    & Appearance $\leftrightarrow$ Background Suppress & 90.8 & 0.46 & 0.14 \\
    & Neutral $\leftrightarrow$ OCR/Text & 88.9 & 0.30 & 0.09 \\    
    \midrule
    \multirow{5}{*}{MM-IMDb}
    & Semantic $\leftrightarrow$ Appearance & 92.1 & 0.58 & 0.17 \\
    & Semantic $\leftrightarrow$ Background Suppress & 94.3 & 0.43 & 0.13 \\
    & Appearance $\leftrightarrow$ Demographic Suppress & 90.6 & 0.49 & 0.15 \\
    & Semantic $\leftrightarrow$ Demographic Suppress & 93.8 & 0.40 & 0.11 \\
    & Neutral $\leftrightarrow$ Semantic & 88.4 & 0.28 & 0.08 \\
    \midrule
    \multirow{5}{*}{CREMA-D}
    & Semantic $\leftrightarrow$ Appearance & 91.7 & 0.55 & 0.18 \\
    & Semantic $\leftrightarrow$ Motion & 89.9 & 0.46 & 0.16 \\
    & Motion $\leftrightarrow$ Appearance & 88.3 & 0.51 & 0.17 \\
    & Semantic $\leftrightarrow$ Background Suppress & 93.6 & 0.41 & 0.12 \\
    & Neutral $\leftrightarrow$ Semantic & 87.8 & 0.26 & 0.07 \\
    \midrule
    \multirow{5}{*}{RAVDESS}
    & Semantic $\leftrightarrow$ Appearance & 91.1 & 0.53 & 0.17 \\
    & Semantic $\leftrightarrow$ Motion & 89.4 & 0.44 & 0.15 \\
    & Motion $\leftrightarrow$ Appearance & 87.9 & 0.49 & 0.16 \\
    & Semantic $\leftrightarrow$ Background Suppress & 92.8 & 0.39 & 0.11 \\
    & Neutral $\leftrightarrow$ Semantic & 87.1 & 0.24 & 0.07 \\
    \midrule
    \multirow{5}{*}{Flickr30K}
    & Semantic $\leftrightarrow$ Appearance & 0.84 & 0.63 & 0.20 \\
    & Semantic $\leftrightarrow$ Appearance Suppress & 0.88 & 0.51 & 0.15 \\
    & Appearance $\leftrightarrow$ Background Suppress & 0.81 & 0.56 & 0.18 \\
    & Semantic $\leftrightarrow$ Background Suppress & 0.90 & 0.44 & 0.12 \\
    & Neutral $\leftrightarrow$ Semantic & 0.79 & 0.29 & 0.08 \\
    \bottomrule
    \end{tabular}
\end{table*}

\begin{figure}[!h]
\centering
\includegraphics[width=\columnwidth]{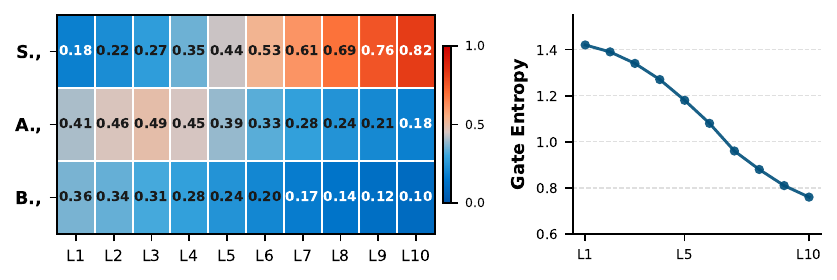}
\caption{
Layer-wise pathway modulation and gate entropy on Flickr30K.
Gate entropy denotes the summed Bernoulli entropy across independent sigmoid gates.
The heatmap shows Semantic, Appearance, and Background pathway activations.
Semantic pathway activation increases in deeper layers, while gate entropy progressively decreases, indicating increasingly selective pathway activation across decoding layers.
}
\vspace{-0.7em}
\label{fig:layerwise_routing}
\end{figure}

\section{Layer-wise Evidence Routing}
\label{app:layerwise}

We next analyze how instruction-conditioned pathway modulation evolves across transformer depth during autoregressive generation.
Figure~\ref{fig:layerwise_routing} shows layer-wise pathway activations on Flickr30K.
Earlier layers exhibit stronger appearance-related activation, whereas later layers increasingly emphasize semantic pathways.
Intermediate layers show more balanced activation patterns, suggesting progressive integration of multimodal evidence before stronger semantic emphasis in deeper layers.
In parallel, gate entropy steadily decreases across layers, indicating increasingly selective pathway activation during generation.
Overall, these results show that GUIDE produces structured instruction-conditioned modulation across transformer depth rather than isolated changes at a single layer.


\subsection{Generalization to Paraphrased and Compositional Instructions}

\begin{figure}[!h]
   \centering
   \includegraphics[width=\columnwidth]{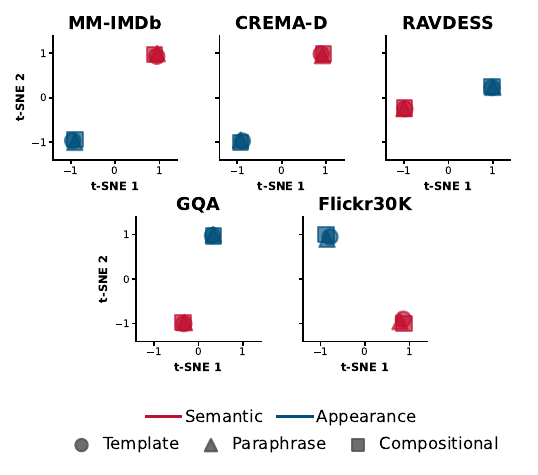}
   \caption{t-SNE of Reliance Sensitivity (RS) vectors for semantic-intent (red) and appearance-intent (blue) instructions across benchmarks and instruction types.}
   \label{fig:tsne_row2_3}
   \vspace{-0.5em}
\end{figure}

To evaluate whether GUIDE generalizes beyond fixed instruction templates, we project Reliance Sensitivity (RS) vectors from template-based, paraphrased, and compositional instruction variants into a 2D space using t-SNE.
As shown in Figure~\ref{fig:tsne_row2_3}, semantic-intent and appearance-intent instructions form consistently separated clusters across benchmarks.
Importantly, this separation remains stable across template, paraphrased, and compositional variants, suggesting that GUIDE captures instruction-conditioned evidence preferences beyond fixed lexical formulations.
These results further support that GUIDE generalizes instruction-conditioned evidence modulation to semantically equivalent and compositionally novel instructions.

\vspace{-0.8em}
\section{Full Ablation Results}
\label{app:full_ablation}
\vspace{-0.5em}

Table~\ref{tab:full_ablation} reports full ablation results across all multimodal reasoning, classification, and generation benchmarks.
Consistent trends are observed across datasets:
removing taxonomy-aligned grouped pathways substantially reduces pathway modulation and changes in functional reliance, while replacing dynamic instruction-conditioned gates with static routing also decreases perturbation robustness.
Prompt-only control consistently exhibits the weakest evidence-modulation behavior, further indicating that grouped adaptation and dynamic gating contribute substantially beyond prompt-only control.
In the no-grouping variant, the taxonomy-aligned branches are replaced by a single shared LoRA branch with an instruction-conditioned scalar gate ($C=1$).

\begin{table}[t]
\centering
\small
\setlength{\tabcolsep}{5pt}
\renewcommand{\arraystretch}{0.95}
\caption{
Ablation study on grouped routing and dynamic gating.
Removing taxonomy-aligned grouped pathways or replacing dynamic instruction-conditioned gates reduces pathway modulation (GMR), changes in functional reliance ($\Delta$RS), and perturbation robustness (Robust.) across benchmarks.
Robust. denotes counterfactual prediction stability under taxonomy-aligned perturbations.
}
\label{tab:full_ablation}

\begin{tabular}{llccc}
\toprule
\textbf{Dataset} & \textbf{Variant} & \textbf{GMR}$\uparrow$ & \textbf{$\Delta$RS}$\uparrow$ & \textbf{Robust.}$\uparrow$ \\
\midrule

\multirow{4}{*}{GQA}
& Full & \textbf{0.61} & \textbf{0.19} & \textbf{0.93} \\
& No grouped & 0.22 & 0.07 & 0.81 \\
& Static gates & 0.38 & 0.10 & 0.86 \\
& Prompt-only & 0.18 & 0.06 & 0.79 \\

\midrule

\multirow{4}{*}{TextVQA}
& Full & \textbf{0.57} & \textbf{0.18} & \textbf{0.92} \\
& No grouped & 0.21 & 0.06 & 0.80 \\
& Static gates & 0.36 & 0.09 & 0.85 \\
& Prompt-only & 0.17 & 0.05 & 0.78 \\

\midrule

\multirow{4}{*}{MM-IMDb}
& Full & \textbf{0.58} & \textbf{0.17} & \textbf{0.92} \\
& No grouped & 0.20 & 0.06 & 0.80 \\
& Static gates & 0.35 & 0.09 & 0.85 \\
& Prompt-only & 0.17 & 0.05 & 0.78 \\

\midrule

\multirow{4}{*}{CREMA-D}
& Full & \textbf{0.55} & \textbf{0.18} & \textbf{0.91} \\
& No grouped & 0.19 & 0.06 & 0.79 \\
& Static gates & 0.33 & 0.09 & 0.84 \\
& Prompt-only & 0.16 & 0.05 & 0.77 \\

\midrule

\multirow{4}{*}{RAVDESS}
& Full & \textbf{0.53} & \textbf{0.17} & \textbf{0.91} \\
& No grouped & 0.18 & 0.06 & 0.78 \\
& Static gates & 0.31 & 0.08 & 0.83 \\
& Prompt-only & 0.15 & 0.05 & 0.76 \\

\midrule

\multirow{4}{*}{Flickr30K}
& Full & \textbf{0.63} & \textbf{0.20} & \textbf{0.84} \\
& No grouped & 0.23 & 0.07 & 0.73 \\
& Static gates & 0.39 & 0.11 & 0.78 \\
& Prompt-only & 0.19 & 0.06 & 0.71 \\

\bottomrule
\end{tabular}
\vspace{-0.8em}
\end{table}

\section{Additional Evidence Routing Visualizations}
\label{app:additional_heatmaps}
\begin{figure}[p]
   \centering
   \includegraphics[width=\columnwidth]{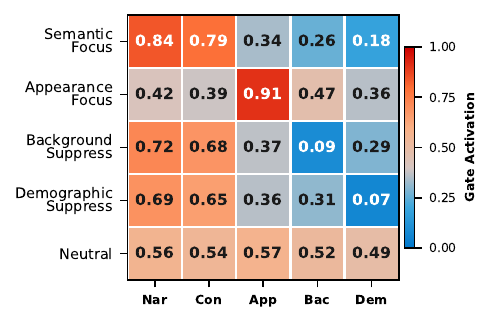}
   \caption{Gate activation heatmap on MM-IMDb under different instruction conditions. X-labels denote \textbf{Nar}rative, \textbf{Con}ceptual, \textbf{App}earance, \textbf{Bac}kground, \textbf{Dem}ographic respectively.}
   \label{fig:heatmap_mmimdb}
\end{figure}

\begin{figure}[p]
   \centering
   \includegraphics[width=\columnwidth]{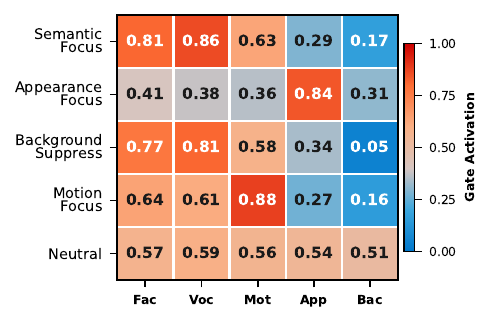}
   \caption{Gate Activation Heatmap on CREMA-D under different instruction conditions. X-labels denote \textbf{Fac}ial Expression, \textbf{Voc}al Emotion, \textbf{Mot}ion, \textbf{App}earance, \textbf{Bac}kground respectively.}
   \label{fig:heatmap_cremad}
\end{figure}

\begin{figure}[p]
   \centering
   \includegraphics[width=\columnwidth]{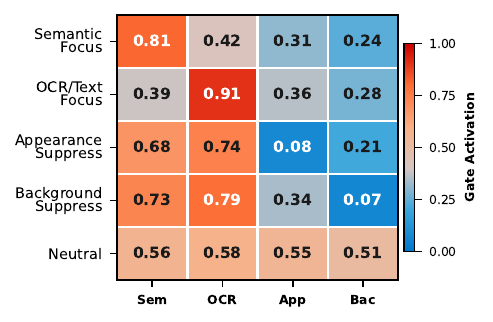}
   \caption{Gate Activation Heatmap on TextVQA under different instruction conditions. X-labels denote \textbf{Sem}antic, \textbf{OCR}/Text, \textbf{App}earance, \textbf{Bac}kground respectively}
   \label{fig:heatmap_textvqa}
\end{figure}

\begin{figure}[p]
   \centering
   \includegraphics[width=\columnwidth]{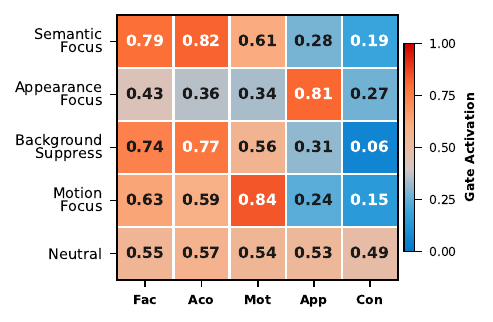}
   \caption{Gate Activation Heatmap on RAVDESS under different instruction conditions. X-labels denote \textbf{Fac}ial Emotion, \textbf{Aco}ustic Tone, \textbf{Mot}ion, \textbf{App}earance, \textbf{Con}text respectively.}
   \label{fig:heatmap_ravdess}
\end{figure}

\begin{figure}[p]
   \centering
   \includegraphics[width=\columnwidth]{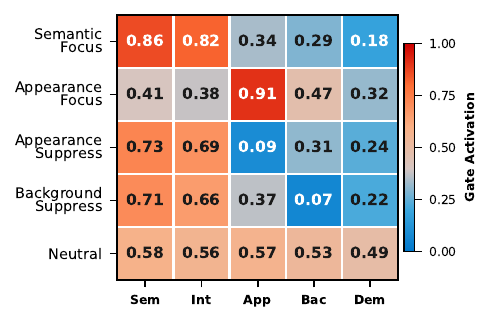}
   \caption{Gate Activation Heatmap on Flickr30K under different instruction conditions. X-labels denote \textbf{Sem}antic, \textbf{Int}eraction, \textbf{App}earance, \textbf{Bac}kground, \textbf{Dem}ographic respectively}
   \label{fig:heatmap_flickr}
\end{figure}

Figures~\ref{fig:heatmap_mmimdb}--\ref{fig:heatmap_flickr} provide additional gate-activation visualizations across multimodal reasoning, classification, and generation benchmarks.
Consistent with the main GQA analysis, instruction-conditioned gating produces structured modulation of pathway activations aligned with the intended evidence preference.
Semantic-focused instructions increase activation of semantic and interaction-related routing groups, whereas appearance-focused instructions increase activation of appearance-related groups.
Suppression-oriented instructions reduce activation of the targeted routing groups while preserving activation of other task-relevant pathways.
These patterns are observed across benchmarks with different modalities and task structures.

\begin{figure}[p]
   \centering
   \includegraphics[width=\columnwidth]{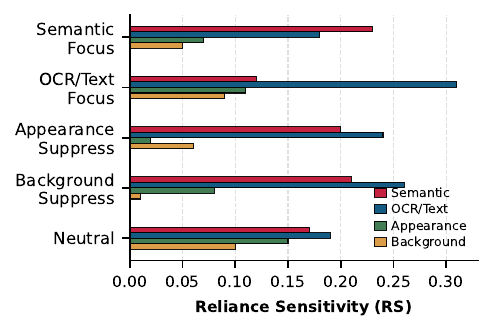}
   \caption{Reliance Sensitivity (RS) across evidence pathways under different instruction conditions on TextVQA. Higher RS indicates greater functional dependence on the corresponding evidence pathway.}
   \label{fig:grouped_barplot_textvqa}
\end{figure}
\begin{figure}[t]
   \centering
   \includegraphics[width=\columnwidth]{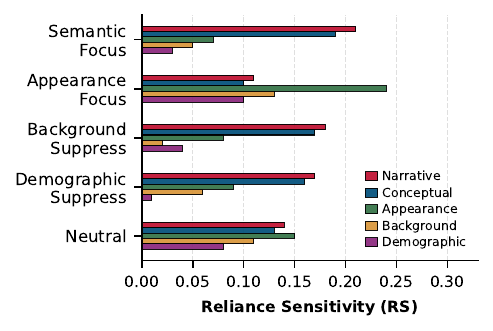}
   \caption{Reliance Sensitivity (RS) across evidence pathways under different instruction conditions on MM-IMDb. Higher RS indicates greater functional dependence on the corresponding evidence pathway.}
   \label{fig:grouped_barplot_mmimdb}
\end{figure}
\begin{figure}[t]
   \centering
   \includegraphics[width=\columnwidth]{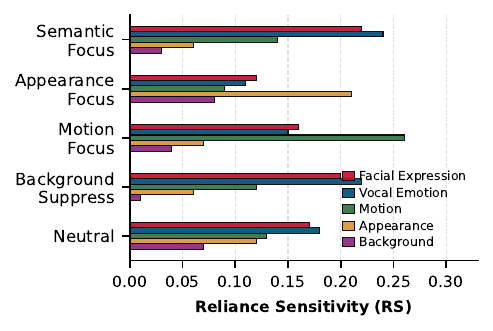}
   \caption{Reliance Sensitivity (RS) across evidence pathways under different instruction conditions on CREMA-D. Higher RS indicates greater functional dependence on the corresponding evidence pathway.}
   \label{fig:grouped_barplot_cremad}
\end{figure}

\section{Additional Functional Reliance Results}
\label{app:additional_rs}

\begin{figure}[!t]
   \centering
   \includegraphics[width=\columnwidth]{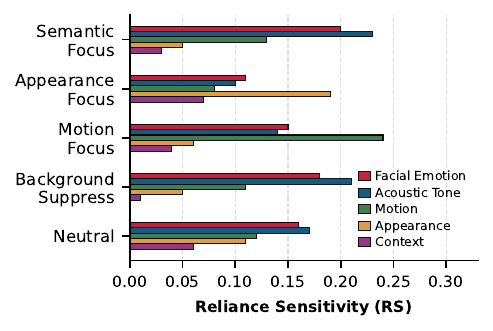}
   \caption{Reliance Sensitivity (RS) across evidence pathways under different instruction conditions on RAVDESS. Higher RS indicates greater functional dependence on the corresponding evidence pathway.}
   \label{fig:grouped_barplot_ravdess}
\end{figure}

\begin{figure}[!t]
   \centering
   \includegraphics[width=\columnwidth]{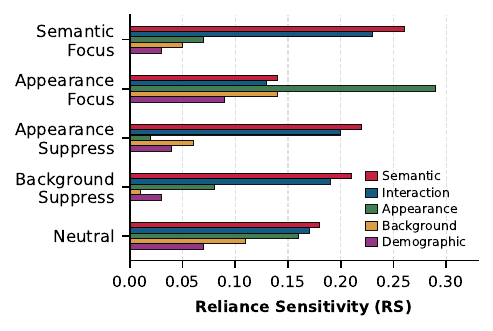}
   \caption{Reliance Sensitivity (RS) across evidence pathways under different instruction conditions on Flickr30K. Higher RS indicates greater functional dependence on the corresponding evidence pathway.}
   \label{fig:grouped_barplot_flickr}
\end{figure}

Figures~\ref{fig:grouped_barplot_textvqa}--\ref{fig:grouped_barplot_flickr} report additional Reliance Sensitivity (RS) analyses across multimodal reasoning, classification, and generation benchmarks.
Consistent with the main GQA results, different instructions produce systematic shifts in functional dependence across evidence pathways.
Semantic-focused instructions increase sensitivity to semantic and interaction-related pathways, whereas appearance-focused instructions increase reliance on appearance-associated evidence.
Suppression-oriented instructions reduce functional dependence on the targeted evidence category while preserving sensitivity to other task-relevant pathways.
Together with the gate-activation results, these findings indicate that GUIDE modulates not only pathway activations but also the evidence that functionally influences model predictions.


\vspace{-1em}
\section{Additional Perturbation Stability Analyses}
\label{app:perturbation_stability}
\begin{figure*}[!h]
   \includegraphics[width=0.32\textwidth]{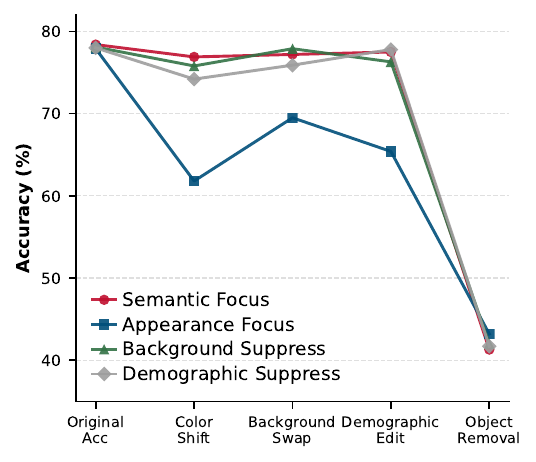}
   \hspace{-0.9em}
   \centering
   \includegraphics[width=0.32\textwidth]{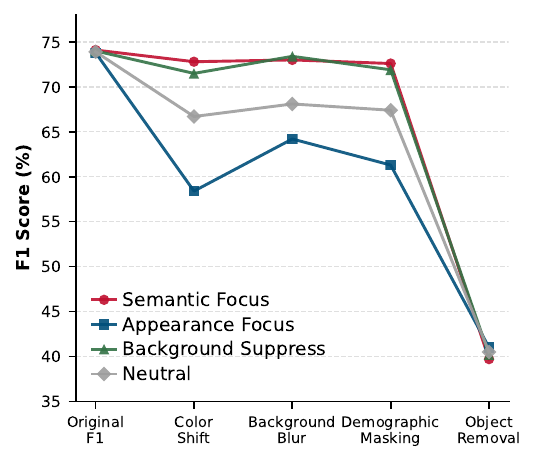}
   \hspace{-0.9em}
   \centering
   \includegraphics[width=0.32\textwidth]{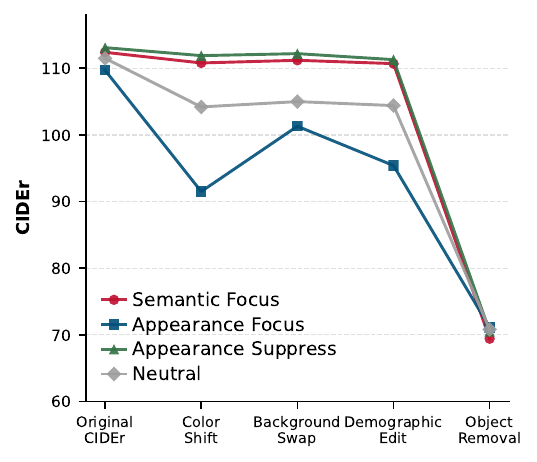}
   \caption{Perturbation-stability results on GQA, MM-IMDb, and Flickr30K under different instruction conditions and evidence perturbations.}
   \label{fig:perturbation_stability}
\end{figure*}

Figure~\ref{fig:perturbation_stability} reports additional perturbation-stability analyses on GQA, MM-IMDb, and Flickr30K.
Consistent with the main GQA results, semantic-focused and suppression-oriented instructions remain comparatively stable under appearance- and context-related perturbations, whereas appearance-focused instructions exhibit larger degradation under appearance-related perturbations.
Across the evaluated datasets, perturbations to core semantic evidence produce the largest performance degradation, indicating that task-relevant semantic evidence remains important despite instruction-conditioned modulation of auxiliary pathways.

\begin{table*}[!h]
\centering
\caption{Dataset Statistics.}
\label{tab:dataset_statistics}
\begin{adjustbox}{max width =\textwidth}
\small
\centering
\begin{tabularx}{\textwidth}{
  >{\raggedright\arraybackslash}p{1.2cm}
  >{\raggedright\arraybackslash}p{1.5cm}
  >{\centering\arraybackslash}p{2.5cm}
  >{\centering\arraybackslash}p{1.0cm}
  >{\centering\arraybackslash}p{1.0cm}
  >{\centering\arraybackslash}p{1.0cm}
  >{\raggedright\arraybackslash}X
}
\toprule
\textbf{Dataset} & \textbf{Task} & \textbf{Modality} & \textbf{Train} & \textbf{Val} & \textbf{Test} & \textbf{High-Level Evidence Families} \\
\midrule
GQA & Reasoning & Vision--Language & 943,000 & 132,062 & 12,578 & Semantic, Appearance, Background, Demographic \\
TextVQA & Reasoning & Vision--Language &34,602 & 5,000 & 5,734 & Semantic, OCR/Text, Appearance, Background \\
MM-IMDb & Classification & Vision--Language & 2,142 & 1,071 & 1,072 & Semantic, Appearance, Background, Demographic \\
CREMA-D & Classification & Audio--Visual & 5,152 & 736 & 1,472 & Semantic, Motion, Appearance, Background \\
RAVDESS & Classification & Audio--Visual & 1,470 & 490 & 491 & Semantic, Motion, Appearance, Background \\
Flickr30K & Captioning & Vision--Language & 28,998 & 1,024 & 992 & Semantic, Appearance, Background \\
\bottomrule
\end{tabularx}
\end{adjustbox}
\end{table*}

\section{Implementation Details}
\label{app:implementation_details}

Table~\ref{tab:dataset_statistics} summarizes the datasets, task types, modalities, dataset splits, and high-level evidence families used in our experiments.
These high-level families provide a shared descriptive taxonomy across benchmarks and should be distinguished from the operational routing groups used by GUIDE.
Operational groups may refine a high-level family into multiple task-specific components.
For example, the semantic family in GQA is implemented using
separate object-, relation-, and action-related routing groups,
whereas the appearance family is implemented using an
attribute-related routing group.
Consequently, the number of operational groups reported in Table~\ref{tab:hyperparameters} may exceed the number of high-level families listed in Table~\ref{tab:dataset_statistics}.

Across tasks, the shared high-level taxonomy includes semantic, appearance, background, motion, OCR/text, and demographic evidence when applicable.
The operational interpretation of these families depends on the modality and benchmark.
For example, semantic routing in GQA captures object, action,
and relational information, whereas appearance routing captures
attribute-related information. CREMA-D and RAVDESS use
task-specific facial, acoustic, and motion-related groups.
OCR/text routing is used for TextVQA, while motion-related routing is used primarily for audio--visual emotion-recognition tasks.

This hierarchical organization provides a common evidence taxonomy while allowing GUIDE to use task-appropriate operational routing groups.

\subsection{Pretrained Multimodal Backbones}
\label{app:backbones}
We evaluate GUIDE across multiple pretrained multimodal foundation models spanning both vision--language and audio--visual settings.

For vision--language benchmarks (GQA, Flickr30K, TextVQA, and MM-IMDb), we instantiate GUIDE on top of Qwen3-VL-30B, Qwen3.5-VL-37B, and InternVL3.5-38B.
For audio--visual benchmarks (CREMA-D and RAVDESS), we use Qwen3-Omni-30B and Ola-7B as multimodal backbones.

These models serve as the underlying multimodal reasoning architectures on top of which grouped adaptation and instruction-conditioned routing are applied.
We additionally observe that the overall evidence-routing behavior and reliance modulation patterns remain qualitatively consistent across different pretrained multimodal backbones.

\begin{table*}[t]
\centering
\small
\caption{
Training hyperparameters used for GUIDE across multimodal benchmarks.
The number of pathway groups is determined by the dataset-specific evidence taxonomy.
}
\label{tab:hyperparameters}
\setlength{\tabcolsep}{4pt}
\renewcommand{\arraystretch}{0.95}

\begin{tabular}{lccccccc}
\toprule
\textbf{Dataset} & \textbf{Backbone} & \textbf{Groups} & \textbf{Rank} & \textbf{LR} & \textbf{Batch} & \textbf{Epochs} & \textbf{$\lambda_{\mathrm{gate}}$} \\
\midrule

GQA 
& Qwen3-VL-30B 
& 6 
& 8 
& $2\times10^{-4}$ 
& 32 
& 5 
& 0.01 \\

TextVQA 
& Qwen3-VL-30B 
& 4 
& 8 
& $2\times10^{-4}$ 
& 32 
& 5 
& 0.01 \\

MM-IMDb 
& InternVL3.5-38B 
& 5 
& 8 
& $1\times10^{-4}$ 
& 64 
& 10 
& 0.01 \\

CREMA-D 
& Qwen3-Omni-30B 
& 5 
& 8 
& $1\times10^{-4}$ 
& 32 
& 10 
& 0.01 \\

RAVDESS 
& Ola-7B 
& 5 
& 8 
& $1\times10^{-4}$ 
& 32 
& 10 
& 0.01 \\

Flickr30K 
& Qwen3.5-VL-37B 
& 5 
& 8 
& $2\times10^{-4}$ 
& 32 
& 5 
& 0.01 \\

\bottomrule
\end{tabular}
\vspace{-0.5em}
\end{table*}

\section{Computational Experiments}
All experiments are conducted using grouped parameter-efficient adaptation on top of frozen pretrained multimodal backbones.
Training and evaluation are performed using standard GPU-based distributed training environments.
Because GUIDE modifies only lightweight grouped LoRA parameters and gating modules while keeping backbone parameters frozen, the computational overhead remains substantially smaller than full model finetuning.
Detailed hyperparameters, backbone configurations, and implementation details are provided in the appendix.

\subsection{Training Details}
\label{app:training_details}
The pretrained multimodal backbone remains frozen during training, and only the grouped LoRA parameters and instruction-conditioned gating network are updated.
We use AdamW with learning rates selected from
$\{1\times10^{-4},2\times10^{-4}\}$ based on validation performance.
The LoRA rank is fixed to 8 across experiments.
GUIDE uses independent sigmoid gates:
each routing-group activation is computed independently and lies in $[0,1]$, without normalization across groups.
Models are trained using mixed precision with batch sizes between 32 and 64, depending on the benchmark and backbone size.
The gate-regularization coefficient $\lambda_{\mathrm{gate}}$ is set to 0.01 across experiments.
For the compositional-routing analysis, we use a separate trade-off coefficient $\alpha\in[0,1]$ to interpolate the relative emphasis of semantic- and appearance-focused instruction components during inference.
All training hyperparameters are selected using the validation split and kept fixed across instruction variants within each benchmark.

\subsection{License for Artifacts}
All experiments are conducted using publicly available datasets and pretrained multimodal foundation models subject to their respective licenses and terms of use.
We release only the code necessary to reproduce the GUIDE routing framework, training procedures, and evaluation protocols.
Users are responsible for ensuring compliance with the licenses and usage restrictions associated with the underlying datasets and pretrained backbones.

\begin{table}[h]
\centering
\caption{Human Evaluation Results. Each field is rated on a Likert scale of 1--4.}
\small
\begin{tabular}{lcc}
\toprule
\textbf{Condition} & \textbf{Alignment} & \textbf{Preservation} \\
\midrule
Semantic Focus & 3.76 & 3.63 \\
Appearance Focus & 3.82 & 3.67 \\
Background Suppress & 3.59 & 3.47 \\
\midrule
\textbf{Grand Total} & \textbf{3.72} & \textbf{3.59} \\
\bottomrule
\end{tabular}
\label{tab:human_eval}
\end{table}
\vspace{-0.5em}
\section{Human Evaluation}

\subsection{Evaluation Setup}
We conducted a human evaluation with eleven annotators: one undergraduate student, one graduate assistant, four Master's students, three PhD students, and two postdoctoral researchers. All annotators had prior experience with multimodal or language-model evaluation. Participation was voluntary and uncompensated, and annotators were informed of the purpose of the study and the intended use of their responses before providing consent.
We randomly sampled 50 Flickr30K examples across the evaluated instruction conditions.
For each example, annotators were shown the source image, instruction, and generated caption and rated
(i) instruction alignment and
(ii) semantic preservation
using a four-point Likert scale.

\subsection{Evaluation Results}
We conducted a human evaluation study on Flickr30K to assess whether instruction-conditioned evidence control remains semantically coherent during generation.
Annotators were given an image, instruction, and generated caption, and asked to rate:
(i) instruction alignment and
(ii) semantic preservation on a 4-point scale.
The evaluation was conducted on 50 randomly sampled examples across different instruction conditions.
Table~\ref{tab:human_eval} reports the average human ratings.
All instruction conditions achieve high alignment and preservation scores, indicating that GUIDE produces captions that both follow the intended evidence preference and preserve overall scene semantics.
Importantly, semantic preservation remains consistently high despite substantial differences in the routing of instruction-conditioned evidence.
\vspace{-0.6em}
\section{Additional Qualitative Examples}
\label{app:qualitative}

We provide additional qualitative examples illustrating how GUIDE changes instruction-conditioned evidence emphasis while preserving the underlying input and task.
Figures~\ref{fig:gqa_examples} and~\ref{fig:textvqa_examples} show representative GQA and TextVQA examples under baseline, semantic-focused, and appearance-focused instructions.

Without an explicit evidence preference, the model exhibits comparatively mixed activation across semantic- and appearance-related routing groups.
Under semantic-focused instructions, GUIDE increases activation associated with task-relevant semantic structure.
For GQA, this includes actions, object interactions, attributes, and relational cues relevant to answering the question.
For TextVQA, semantic-focused instructions emphasize contextual and relational information, while OCR-focused instructions, when applied, emphasize localized textual evidence.

Appearance-focused instructions instead increase activation associated with visual attributes, layout, style, clothing, and other appearance-related cues.
These changes illustrate that identical multimodal inputs can produce different pathway-activation patterns under contrasting evidence instructions.

The examples are consistent with the quantitative GMR and RS analyses: gate visualizations show changes in pathway activation, while the perturbation-based RS results establish corresponding changes in functional reliance.
Overall, the qualitative examples illustrate how GUIDE modulates evidence processing across reasoning and generation settings.


\end{document}